\documentclass[conference]{IEEEtran}
\IEEEoverridecommandlockouts
\usepackage{adjustbox} 
\usepackage{amsmath,amssymb,amsfonts}
\usepackage{algpseudocode}
\usepackage{algorithm}

\usepackage{multirow}

\usepackage{graphicx}
\usepackage{textcomp}
\usepackage{float}

\usepackage{cite}
\usepackage{hyperref}
\hypersetup{
    colorlinks=true,
    linkcolor=blue,
    filecolor=magenta,      
    urlcolor=cyan,
    pdftitle={Overleaf Example},
    }

\usepackage{adjustbox} 

\IEEEoverridecommandlockouts

\usepackage{tikz}
\usepackage{textcomp}
\usepackage{hyperref}
\usepackage{lipsum}

\newcommand\copyrighttext{%
  \footnotesize \textcopyright 2021 IEEE. Personal use of this material is permitted. Permission
from IEEE must be obtained for all other uses, in any current or future
media, including reprinting/republishing this material for advertising or
promotional purposes, creating new collective works, for resale or
redistribution to servers or lists, or reuse of any copyrighted
component of this work in other works.}
\newcommand\copyrightnotice{%
\begin{tikzpicture}[remember picture,overlay]
\node[anchor=south,yshift=10pt] at (current page.south) {\fbox{\parbox{\dimexpr\textwidth-\fboxsep-\fboxrule\relax}{\copyrighttext}}};
\end{tikzpicture}%
}
\makeatletter
\def\maxwidth{%
  \ifdim\Gin@nat@width>\linewidth
    \linewidth
  \else
    \Gin@nat@width
  \fi
}
\makeatother

\begin{document}

\title{HM-Net: A Regression Network\\ for Object Center Detection and Tracking\\ on Wide Area Motion Imagery}

\author{
\IEEEauthorblockN{1\textsuperscript{st} Hakk{\i} Motorcu}\
\IEEEauthorblockA{
\textit{Electrical and Electronics}\\ \textit{ Engineering Department} \\
\textit{Ozyegin University}\\
Istanbul, Turkey \\
hakki.motorcu@ozu.edu.tr}
\and

\IEEEauthorblockN{2\textsuperscript{nd} Hasan F. Ates}
\IEEEauthorblockA{\textit{School of Engineering}\\ \textit{ and Natural Sciences} \\
\textit{Istanbul Medipol University}\\
Istanbul, Turkey \\
hfates@medipol.edu.tr}

\and
\IEEEauthorblockN{3\textsuperscript{rd} H. Fatih Ugurdag}
\IEEEauthorblockA{\textit{Electrical and Electronics}\\ \textit{ Engineering Department} \\
\textit{Ozyegin University}\\
Istanbul, Turkey \\
fatih.ugurdag@ozyegin.edu.tr}

\and
\IEEEauthorblockN{4\textsuperscript{th} Bahadir Gunturk}
\IEEEauthorblockA{\textit{School of Engineering}\\ \textit{ and Natural Sciences} \\
\textit{Istanbul Medipol University}\\
Istanbul, Turkey \\
bkgunturk@medipol.edu.tr}

}
\maketitle
 \copyrightnotice

\begin{abstract}
  Wide Area Motion Imagery (WAMI) yields high resolution images with a large number of extremely small objects. Target objects have large spatial displacements throughout consecutive frames. This nature of WAMI images makes object tracking and detection challenging. In this paper, we present our deep neural network-based combined object detection and tracking model, namely, Heat Map Network (HM-Net). HM-Net  is significantly faster than state-of-the-art frame differencing and background subtraction-based methods, without compromising detection and tracking performances. 
  HM-Net follows object center-based joint detection and tracking paradigm. Simple heat map-based predictions support unlimited number of simultaneous detections. The proposed method uses two consecutive frames and the object detection heat map obtained from the previous frame as input, which helps HM-Net monitor spatio-temporal changes between frames and keep track of previously predicted objects. Although reuse of prior object detection heat map acts as a vital feedback-based memory element, it can lead to unintended surge of false positive detections. To increase robustness of the method against false positives and to eliminate low confidence detections, HM-Net employs novel feedback filters and advanced data augmentations. HM-Net outperforms state-of-the-art WAMI moving object detection and tracking methods on WPAFB dataset with its 96.2\% F1 and 94.4\% mAP detection scores, while achieving 61.8\% mAP tracking score on the same dataset. This performance corresponds to an improvement of 2.1\% for F1, 6.1\% for mAP scores on detection, and 9.5\% for mAP score on tracking over state-of-the-art.  
\end{abstract}


\section{Introduction}

\label{sec:introduction}
Object detection and tracking are broad and active subfields of computer vision. Recent improvements in deep neural networks and hardware capabilities have attracted more attention to these fields. Object detection and tracking applications on Wide Area Motion Imagery (WAMI) have had its share of the attention, and today it still holds its place as a challenging computer vision task. WAMI deals with monitoring a large area, which is several kilometers in diameter, with an airborne sensor consisting of multiple cameras. Captured frames from cameras are stitched to obtain a high-resolution image of the target area. Generally, a registration algorithm is applied to consecutive images to compensate for the camera motion. Object detection and tracking with WAMI have numerous applications in both civilian and military domains \cite{25,31,33,35}. Traffic monitoring, driver behavior analysis, disaster response, and wildlife protection are just some examples in civilian domain. Military domain applications include intelligence-gathering, reconnaissance, border security, and surveillance.  

Unmanned Aerial Vehicles (UAV) and WAMI sensors became more accessible in recent years. As a result of that trend, demand for robust detection and tracking algorithm for wide-area surveillance is constantly increasing.

\begin{figure*}[ht]
\includegraphics[width=\maxwidth]{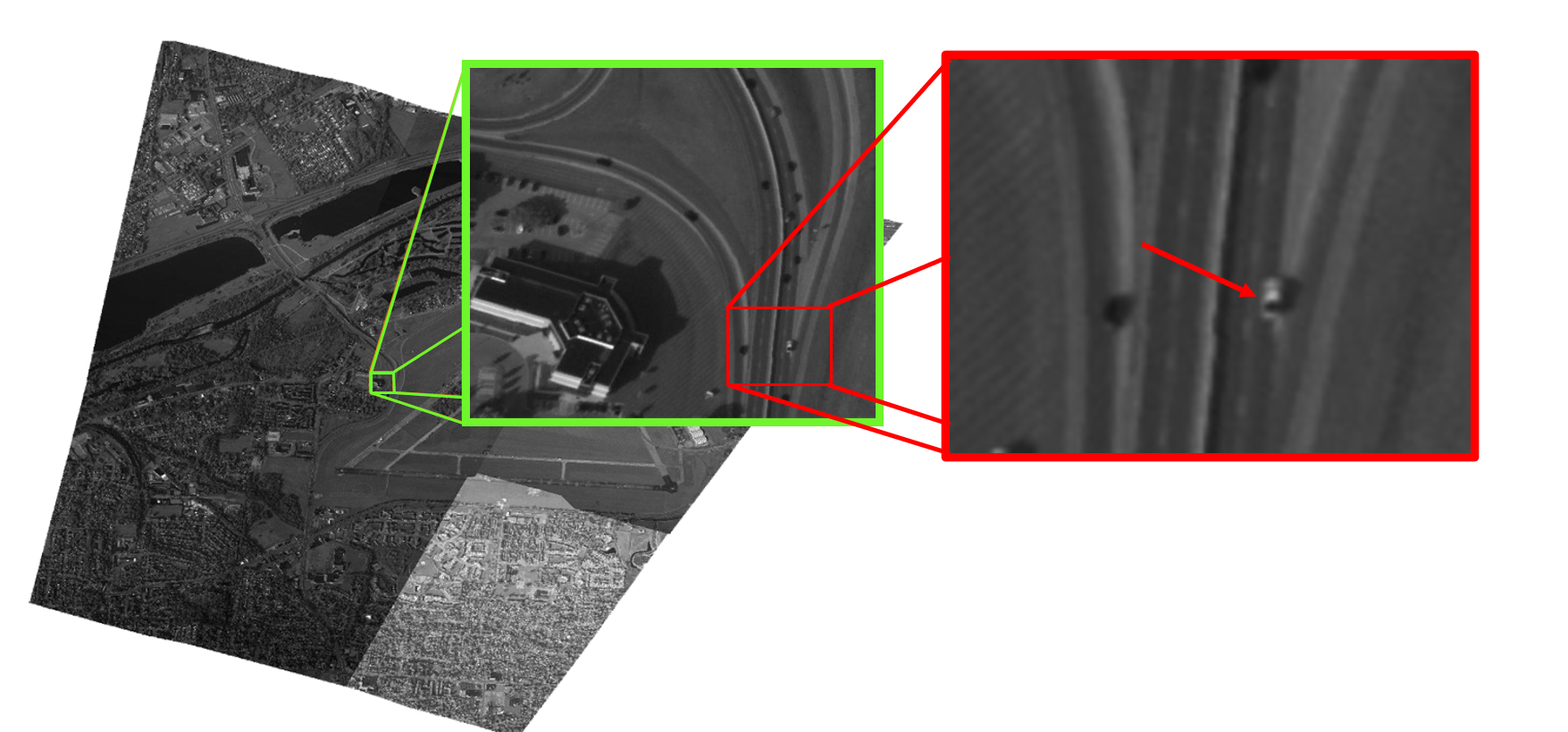}
\caption{A frame from a WAMI footage \cite{36} is magnified to show a vehicle location.}
\label{fig1}
\end{figure*}

Object detection in wide-area images has unique challenges compared to regular object detection \cite{17,25,30}. Although wide-area images have higher resolution, since the area represented by a single image is large, each target object is tiny (see Fig. \ref{fig1}) compared to normal object detection targets \cite{25,30}. Smaller target objects result in fewer features to exploit by the detection algorithm. Usually, WAMI sensors collect grayscale images, and that one channel input yields less information compared to three-channel RGB input. WAMI sensor can capture several thousand objects within a single frame due to high camera point of view, unlike regular video feeds, which are used in regular object detection. Also, ratio of combined pixel area of the objects to total image area is respectively smaller in WAMI. Ground truth information typically contains just center coordinates of objects; other necessary properties for supervised learning like object size and orientation are not provided. From object tracking perspective WAMI introduces large object displacements between consecutive frames and nearly none of the moving objects intersect with their appearances on previous frames. These large displacement vectors are due to slow frame update rate and the high speed of moving vehicles. Overall, a WAMI object tracker should detect and track high number of small objects which carry less appearance information and move very large distances between consecutive frames.

In order to address these existing challenges, we propose a deep neural network-based combined object detector and tracker architecture. Our method represents object locations as a heat map, where each object is represented as a 2D Gaussian placed on its center coordinate. By using center-based object detection we eliminate the need for bounding boxes, which are typically not provided by the WAMI datasets. Our method uses temporal information alongside appearance information to get better detection and tracking results. We use the previous frame alongside the current frame as input to our algorithm. Also, we use a feedback loop that feeds the filtered previous heat map as input. Our method uses deep neural network backbone HM-Net (Heat Map Network), which is specially built and trained for this task. HM-Net employs multiple separate encoders for each input to help model focus on each input independently and then fuses separately extracted features to reduce memory footprint and computational complexity of the model. HM-Net introduces shortcuts between layers and several activation adder blocks to increase overall model connectivity. One-stage nature of the model yields remarkable throughput rate compared to most state-of-the-art algorithms.     

Our model training and feedback procedures are designed to address unique challenges of the WAMI object detection and tracking tasks. For more efficient feedback process and to increase detection performance, elimination of false positives is essential. False positive centers tend to accumulate on heat maps; when the network uses the previous predicted heat map as input, these false detection clusters have tendency to propagate and grow after several feedback cycles. To eliminate false positive propagation during inference, we apply several novel training techniques. In model training we adopt new data augmentation strategies to simulate false positive clusters and false confidence amplifications on feedback heat maps. These measures are taken to make  our model learn the challenging real-life situations on feedback process. Also we use class based selective filtering and rendering algorithm to eliminate potential false positives or low confidence detections during inference. As a result of these novel features we have achieved state-of-the-art performance on both WAMI-based object detection and tracking, and our method is several times faster than recent WAMI detection and tracking algorithms. Our key contributions are:

\begin{itemize}
\item One-stage HM-Net deep neural network architecture for joint object detection and tracking. HM-Net does not need background subtraction or any other preprocessing, is relatively faster than state-of-the-art WAMI detection algorithms and detects both moving and stationary targets with high accuracy.

\item Introduction of Selective Gaussian Reconstruction (SGR) filter. SGR filter is applied during heat map feedback process. SGR aims to reduce false positive propagation error caused by the feedback loop by using two different mechanisms: 1) Class based confidence thresholding to eliminate low confidence detections
2) Confidence amplification for a high confidence detection to avoid potential false positives in its vicinity on the predicted heat map. 

\item  Novel data augmentation techniques for training: Randomized Confidence Reduction (RCR) and Randomized Center Propagation (RCP) are designed for the training process to eliminate vanishing confidence and false positive propagation that is observed during inference.
\end{itemize}

 In Section \ref{sec:relatedwork}  state-of-the-art approaches are analyzed for their strengths and weaknesses. Section \ref{sec:RDT} explains our combined detection and tracking algorithm in detail. Tracking and detection performance of our approach is compared against state-of-the-art models in Section \ref{sec:RAD}. Finally, conclusions and future research are listed in Section \ref{sec:CFW}. 
 
\section{Related work}
\label{sec:relatedwork}
WAMI-based object detection can be classified as spatial, temporal, and spatio-temporal methods according to which information is mostly utilized for detection. Since a complete object tracker has to detect objects for tracking, discussion will be mostly towards the state-of-the-art detection models and their related tracking applications in this section. Regular object detection and tracking tasks are mainly follow objects, which are closer to the are that the camera focuses. Those objects occupy large portions of the image and have more spatial features to exploit during detection and tracking. State-of-the-art detection algorithms \cite{2,3,4,5,6,7,8,9,10} heavily rely on the appearance information of those objects. Since, on WAMI, objects are smaller, less detailed, and occupy less total area on the image, regular detection algorithms are likely to fail in detect target objects. 

\begin{figure*}[ht]
\includegraphics[width=\maxwidth]{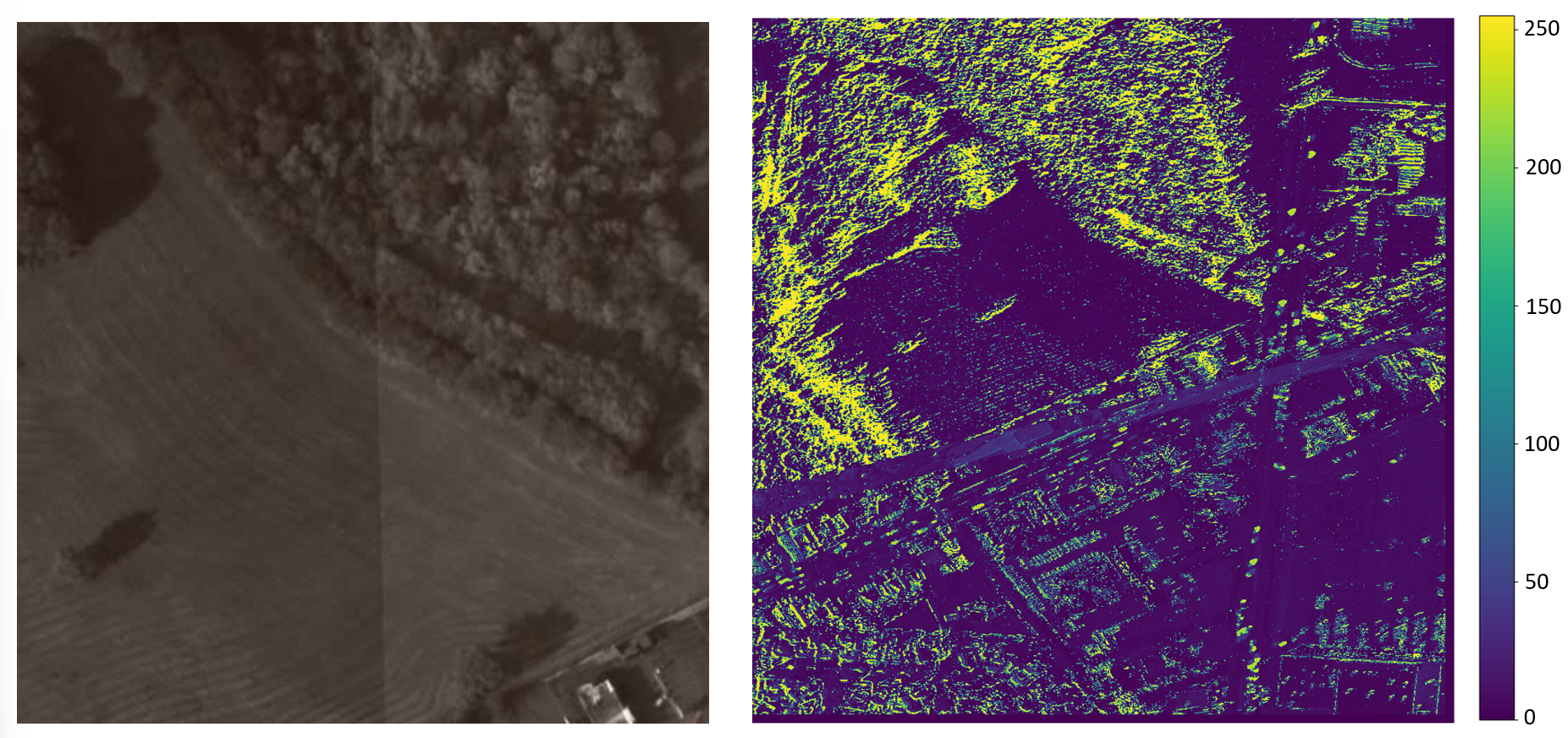}
\caption{A stitching artifact example, which occurs due to different illumination levels of stitched images on the left. Differences between two consecutive registered frames are shown on the right. Error map shows unintended shifts during registration.}  
\label{figK}
\end{figure*}

State-of-the-art single object and  Multi-Object Tracking (MOT) algorithms \cite{11,12,fairmot,14,15} are mostly built on tracking-by-detection paradigm or joint detection and tracking paradigm. Hence, these tracking algorithms inherit problems related to their object detection backbones. In tracking-by-detection approach, most trackers \cite{11,12,14} utilize intersections of the detections between consecutive frames for tracking. Although object re-identification (ID transfer) with intersections yields decent results in high proximity (regular) target objects, which have less displacement throughout time, on WAMI nearly none of the moving objects have any intersection throughout consecutive appearances due to the low frame rate. An alternative paradigm to tracking-by-detection is joint detection and tracking, which is used by \cite{fairmot, 15,1}. In joint detection and tracking, detection model supports tracking algorithm with extra information such as future or previous locations prediction of detected objects to ease target identity transfer. Joint algorithms are more resilient to distortions on inputs and give better results in tracking targets with extreme displacements that yield no appearance overlap.

State-of-the-art WAMI object detectors take advantage of the temporal changes in a video sequence. When  consecutive images are registered (each pixel corresponds to the same location on each image)  moving objects will change pixel values around its location in an ideal scenario. Background subtraction and consecutive frame differencing are main approaches to detect temporal differences, but have their challenges. Background subtraction-based algorithms \cite{16,17,18,19,20,32} use registered video frames to build a background model. From each frame, the algorithm subtracts the modeled background to detect changes and hence finds objects. Background subtraction-based detection can be used solely on moving objects and cannot detect stationary objects. Even slow objects can cause detection problems. Parallax effect, stitching artifacts, illumination variations (see Fig. \ref{figK}) between frames, 
and minor frame registration problems will lead to false detections. Construction of the background model also takes several frames and increases algorithmic complexity and memory utilization. Frame differencing-based methods \cite{21,22,saleemi,xiao,keck} also rely on registered images to detect motion. Unlike background subtraction, frame differencing-based methods rely on fewer consecutive images. These methods use pixel-wise intensity variations between several consecutive frames (generally 2-3). Frame differencing methods suffer from similar problems with background subtraction. Both algorithms need perfect stitching, stable global illumination, and ideal frame registration for successful detections. Any distortion will lead to a significant number of false detections. 

\begin{table*}[!htbp]
\centering
\caption{Summary of WAMI detection and tracking methods in literature.}
\label{tab:RWSS}

\begin{tabular}{|l|l|l|l|l|}
\hline
Methods& Detector &
  Use of Temporal Info. &
  Tracker &
  Limitations \\ \hline
  
 Saleemi {\it et al.} \cite{saleemi} &\begin{tabular}[c]{@{}l@{}}Adaptive consecutive \\frame differencing\end{tabular}&
  \begin{tabular}[c]{@{}l@{}} 2-frame differencing\end{tabular} &
  \begin{tabular}[c]{@{}l@{}}Multiple Hypothesis\\ Tracker (MHT) and  \\  Hungarian  matching\\ algorithm\end{tabular} &
  \begin{tabular}[c]{@{}l@{}}Overall speed   of the \\algorithm is low.  Absence of\\ an explicit 1–1 matching \\ constraint. Can   not detect\\ stationary   vehicles.\end{tabular} \\ \hline
Xiao  {\it et al.} \cite{xiao} &\begin{tabular}[c]{@{}l@{}}Frame differencing and\\ tracking-based target\\ refinement \end{tabular}&
  \begin{tabular}[c]{@{}l@{}} 3-frame differencing\end{tabular} &
  \begin{tabular}[c]{@{}l@{}}The   Graph Matching\\  Framework combined with\\  Vehicle Behavior Modeling\end{tabular} &
  \begin{tabular}[c]{@{}l@{}}Overall   speed of the \\algorithm is low.  Can not \\detect stationary vehicles. \\ Sensitive   to distortions in\\ input images.\end{tabular} \\ \hline
Keck  {\it et al.} \cite{keck} &\begin{tabular}[c]{@{}l@{}}Frame differencing and \\ thresholding\end{tabular}&
  \begin{tabular}[c]{@{}l@{}} 3-frame differencing\end{tabular} &
  \begin{tabular}[c]{@{}l@{}}MHT   with a Kalman filter\\  position/velocity motion \\ model\end{tabular} &
  \begin{tabular}[c]{@{}l@{}}Can   not detect stationary\\ vehicles. Sensitive to distortions\\  in input images.    Struggles \\in highly populated areas.\end{tabular} \\ \hline
Reilly {\it et al.}\cite{32} &\begin{tabular}[c]{@{}l@{}}Background subtraction\end{tabular}&
  \begin{tabular}[c]{@{}l@{}}The median background\\ model is build for \\every 10 frame cycle\end{tabular} &
  \begin{tabular}[c]{@{}l@{}}Bipartite   graph which\\ uses  Hungarian matching\\ algorithm\end{tabular} &
  \begin{tabular}[c]{@{}l@{}}Method   cannot detect\\ stationary vehicles. Sensitive\\ to input distortions.\end{tabular} \\ \hline
Pflugfelder {\it et al.} \cite{30} &\begin{tabular}[c]{@{}l@{}}Modified version of\\ FoveaNet and thresholding\end{tabular}&
  \begin{tabular}[c]{@{}l@{}}Up to 5 prior   frames\\ are used by  CNN model \\for vehicle detection\end{tabular} &
  - &
  \begin{tabular}[c]{@{}l@{}}Method   cannot detect \\stationary  vehicles. Simple \\thresholding based  detection\\   struggles on occluded and\\  close vehicles.\end{tabular} \\ \hline
ClusterNet \cite{25} &\begin{tabular}[c]{@{}l@{}}Two stage CNN architecture\\(FoveaNet and  ClusterNet)\end{tabular}&
  \begin{tabular}[c]{@{}l@{}}Up to 5 prior   frames \\are used by  CNN model\\ for vehicle detection\end{tabular} &
  - &
  \begin{tabular}[c]{@{}l@{}}Two   stage nature of the\\ model increases \\ computational complexity \\and reduces    speed.\end{tabular} \\ \hline
Zhou {\it et al. }\cite{24} &
 \begin{tabular}[c]{@{}l@{}}Background subtraction\\ combined with CNN   \end{tabular} &
  \begin{tabular}[c]{@{}l@{}} Background modeling\\ (5 frame model)\end{tabular} &
  GM-PHD   filtering &
  \begin{tabular}[c]{@{}l@{}}Speed   of the method is\\ vulnerable to  number of proposed\\ detections. Expensive  \\ background subtraction  reduces \\model speed. Method can not \\ detect stationary   vehicles.\end{tabular} \\ \hline
CenterTrack \cite{1} &  \begin{tabular}[c]{@{}l@{}}Modified version of\\ CenterNet\end{tabular}&
  \begin{tabular}[c]{@{}l@{}}Two consecutive frames \\ used as input to CNN \end{tabular} &
  \begin{tabular}[c]{@{}l@{}}Joint   detection and tracking \\ with greedy matching \\ supported by one stage CNN \\ predictions\end{tabular} &
  \begin{tabular}[c]{@{}l@{}}Model is originally designed\\ to work in  regular object \\ detection/tracking tasks.\end{tabular} \\ \hline
\end{tabular}
\end{table*}
Recent approaches for object detection on WAMI (see Table \ref{tab:RWSS}) mostly use motion and appearance information together. This information fusion is generally done with Convolutional Neural Networks (CNN). As discussed above motion-based approaches give false detections due to stitching artifacts, registration errors and sudden pixel illumination changes. To filter false detections \cite{24} utilizes a CNN postprocessing module. Although CNN-based filtering decreases number of false detections, addition of a deep neural network on top of an already computationally expensive background subtraction algorithm decreases overall speed. As discussed above, regular object detection algorithms rely on just spatial information and do not perform well on WAMI. Modified versions of deep object detectors \cite{25,30,26,27,28}  use temporal information to increase overall detection success in both regular detection and WAMI scenarios. These models propose alternative ways to eliminate costly preprocessing methods, i.e. background subtraction and frame differencing. \cite{26} and \cite{28} use 3D CNN and stacked video frames for utilizing temporal information. \cite{27} proposes a CNN that uses separate stacks carrying video frames and optical flow information. \cite{29} proposes object tracking and trajectory calculation boost to object detection by including long short-term memory sub-network and optical flow guided tracking algorithms. ClusterNet \cite{25} uses a two-stage CNN that utilizes FoveaNet \cite{37}. CNN-based detection approach enables ClusterNet \cite {25} to detect stationary targets and moving objects simultaneously.

\begin{figure*}[htp]
\centering
\includegraphics[scale=0.27]{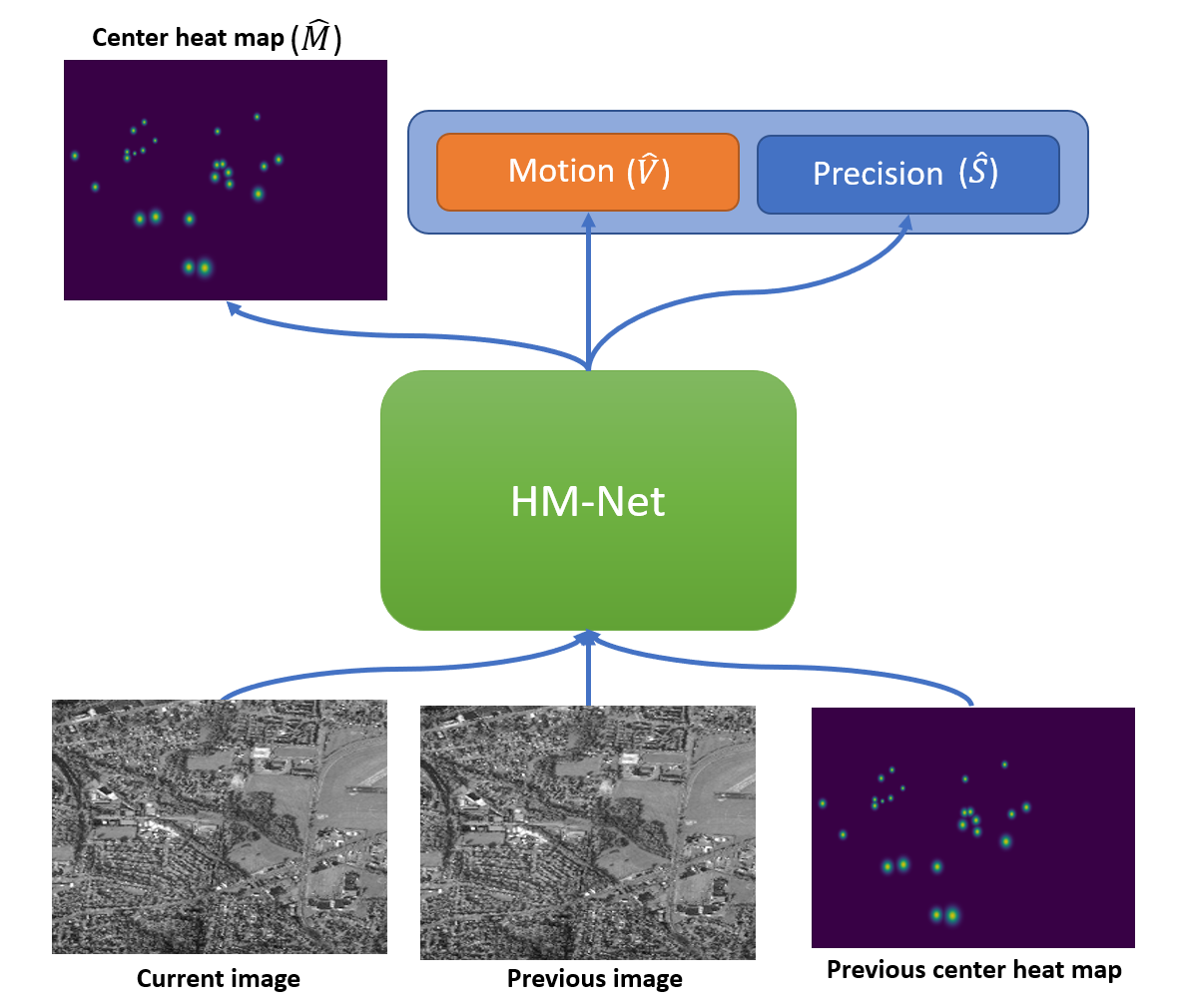}
\caption{Input/Output (I/O) diagram of HM-Net.  }
\label{figIO}
\end{figure*}

\section{Regression Based Detection and Tracking}
\label{sec:RDT}

In this paper, we propose a deep neural network-based joint detection and tracking algorithm for WAMI. Inspired by center-based detection \cite{5,1, 44} and tracking \cite{1} algorithms, our model adopts an anchor-box free approach using a heat map-based object localization. The model uses a one-stage auto-encoder architecture HM-Net as the backbone. Detection and tracking combined throughput rate of the method is about 4.5 fps on a RTX2080Ti 11GB GPU for marked areas of interest (AOIs) in WPAFB dataset \cite{36}. For detecting and tracking moving objects, we split objects into moving vehicle and stationary vehicle classes. Separation of moving and stationary objects enables model to learn certain transitions between classes and reduces false positive detections caused by recently stopped objects. (Solely moving objects are considered during performance evaluation.)

Our method utilizes a center-based joint object detection and tracking paradigm. HM-Net predicts object centers heat map $\hat{M}\in[0,1]^{C \times W \times H}$ ($C$: The number of object classes, $W$: width of input image, $H$: height of input image), motion heat map $\hat{V}\in\mathbb{R}^{2\times W\times H}$, and  subpixel location refinement heat map $\hat{S}\in[0,1]^{2\times W\times H}$ (see Fig. \ref{figIO}). These heat maps hold predictions of object center locations ($\mathbf{m}_i\in \mathbb{R}^2$), motion vectors ($\mathbf{v}_i\in \mathbb{R}^2$), and subpixel refinement of object locations ($\mathbf{s}_i\in \mathbb{R}^2$), respectively. Each heat map shares same width and height dimensions with the input image $I\in \mathbb{R}^{3 \times W \times H}$. Object center points $\mathbf{m}_i$ are represented with 2D Gaussian-shaped peaks. Each predicted peak $\hat{\mathbf{m}}_i=(\hat{x}_i,\hat{y}_i)$ in $M_{c,:,:}$ indicates that there is an object center at coordinates $(\hat{x}_i,\hat{y}_i)$ and it belongs to class $c$. Instead of labeling just center location with full confidence score of one, fitting a Gaussian curve encourages generalization \cite{1}. To render a Gaussian peak around each object center $\mathbf{m}_i$ on heat map $M$, we use the rendering function $R$ given in \ref{eq1}, such that $M_{c,:,:}=R(\lbrace \mathbf{m}_0,\mathbf{m}_1...\mathbf{m}_n\rbrace)$ \cite{42}:

\begin{equation} \label{eq1}
\begin{split}
R_\mathbf{p}(\lbrace \mathbf{m}_0,\mathbf{m}_1...\mathbf{m}_n\rbrace)=\max_i {\exp\left(-\frac{(\mathbf{m}_i-\mathbf{p})^2}{2\sigma^2}\right)}
\end{split}
\end{equation}
where $\mathbf{p}\in \mathbb{R}^2$ is any given position in the heat map, and $\sigma$ is a constant scale parameter that is determined by average object size. 

Local maxima of the predicted heat map $\hat{M}$ yield the integer coordinates of the pixels that hold the object centers. To obtain vital subpixel locations, we use a separate precision head to regress subpixel position of the center ($\mathbf{s}_i\in[0,1]^2$): 

\begin{equation} \label{eq2}
\mathbf{s}_i= \mathbf{m}_i-\lfloor\mathbf{m}_i\rfloor
\end{equation}

\subsection{Detection and Tracking}
\label{sec:RDTa}
Object localization process is guided by the predicted center location heat map $\hat{M}$. As mentioned above, each local maximum $\hat{\mathbf{p}}_i$ of $\hat{M}_{c,:,:}$ corresponds to the center of a detected object with class $c$ and a confidence value $\hat{w}_i=\hat{M}_{c,\hat{\mathbf{p}}_i}$. The center location acquisition from $\hat{M}$ is done within two steps: 1) Non-maxima suppression (NMS): max-pooling operation with window size $15\times15$ is applied to ensure adjacent high confidence pixels of a local maximum will not be mistaken as another local maximum.  2) Class specific threshold $\theta_c$ is applied to eliminate low confidence local maximums. Points that pass thresholding are considered as detected integer object locations. For subpixel center locations HM-Net uses $\hat{\mathbf{s}}_i=\hat{S}_{\hat{\mathbf{p}}_i}$ prediction which yields subpixel coordinates $\hat{x}_i\in[0,1]$, $\hat{y}_i\in[0,1]$ and $\hat{\mathbf{s}}_i=(\hat{x}_i,\hat{y}_i)$. Finally integer and subpixel coordinate estimates are added to yield exact object location predictions: $\hat{\mathbf{m}}_i = \hat{\mathbf{p}}_i + \hat{\mathbf{s}}_i$. 

To track objects through time, we utilize our network to predict 2D motion vectors between consecutive frames using the motion head of the network: $\hat{\mathbf{v}} = \hat{V}_{\hat{\mathbf{p}}}$. Then, we organize all predictions to represent objects during tracking as $\hat{o}=\lbrace \hat{\mathbf{m}_i},\hat{w}_i,\hat{\mathbf{v}}_i,id_i\rbrace^N_{i=0}$, where $id_i$ is the assigned distinct track identity (ID) for each object. HM-Net uses $\hat{o}^{(t-1)}$ and $\hat{o}^{(t)}$ detections for object re-identification. 
From $\hat{o}^{(t)}$ center locations, $\hat{m}_{i}^{(t)}$  and motion vector $\hat{\mathbf{v}_i}^{(t)}$ are utilized to predict potential previous location $\hat{b}_i=\hat{m}_{i}^{(t)} + \hat{\mathbf{v}_i}^{(t)}$ of the object. Detections in $\hat{o}^{(t-1)}$ that are  within an acceptable distance $\kappa$ around $\hat{b}_i$ are marked as potential matches (where $\kappa$ a constant scale parameter that is determined by average object size).  These marked detections in $\hat{o}^{(t-1)}$ are matched with the corresponding candidates in $\hat{o}^{(t)}$ according to their confidence scores ($\hat{w}_i$) by using Hungarian Matching algorithm \cite{48}. Associated objects' tracking IDs are transferred from $\hat{o}^{(t-1)}$ to $\hat{o}^{(t)}$. 

During center matching if a center has no match in the previous frame, we assume that center represents a new object and we generate a new ID for that object. For unmatched prior IDs, we keep such moving objects in a separate set, in case the object have stopped momentarily and classified as stationary or the object confidence is temporarily below the detection threshold. These object tracks could be re-activated in the upcoming frames if a high confidence match is found. Due to the fast motion characteristics of objects in WAMI, this re-identification of a passive track is only possible in the next frame.

\begin{figure}[htb]
\includegraphics[width=\maxwidth]{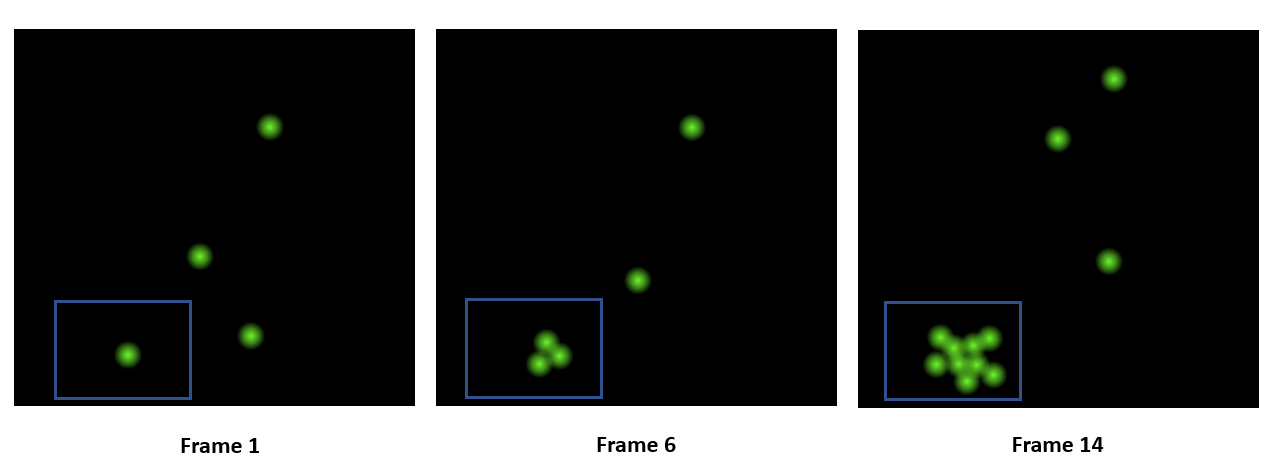}
\caption{A false positive propagation example and its evolution trough time is shown within blue rectangle.}  \label{figFPP}
\end{figure}
Reuse of previous detections via feedback input holds an important place during tracking. Center-based trackers \cite{1,fairmot} exploit the Gaussian-shaped object representation during feedback process. They render all detection information into one channel heat map and use it as feedback. This feedback loop gives the prediction network further temporal information. Although feedback loop adds a short-term memory element to the network, it yields several problems like false-positive error propagation and vanishing or exploding confidence. False positives on a feedback heat map may lead network to generate many less confident false positives that are adjacent to each other. Early experiments revealed that within several iterations, false detection clusters may occupy a significant portion of the output (see Fig. \ref{figFPP}). 

Each detection has confidence score (between 0-1) determined by the local maximums on the predicted heat map. During the feedback process prediction network may amplify or suppress confidence of the detections. An effectively trained network can eliminate potential false positives or false negatives caused by feedback. Efficiency of training depends on the coverage of corner cases like sudden illumination changes within image due to stitching artifacts, minor image registration errors, and problematic areas that can cause ID switches between close objects like junctions etc. That is why countermeasures are taken against these challenges  not only in feedback process but also during the training as well. Training tecnniques will be discussed in detail on the model training section.

\begin{figure}[htb]
\includegraphics[width=\maxwidth]{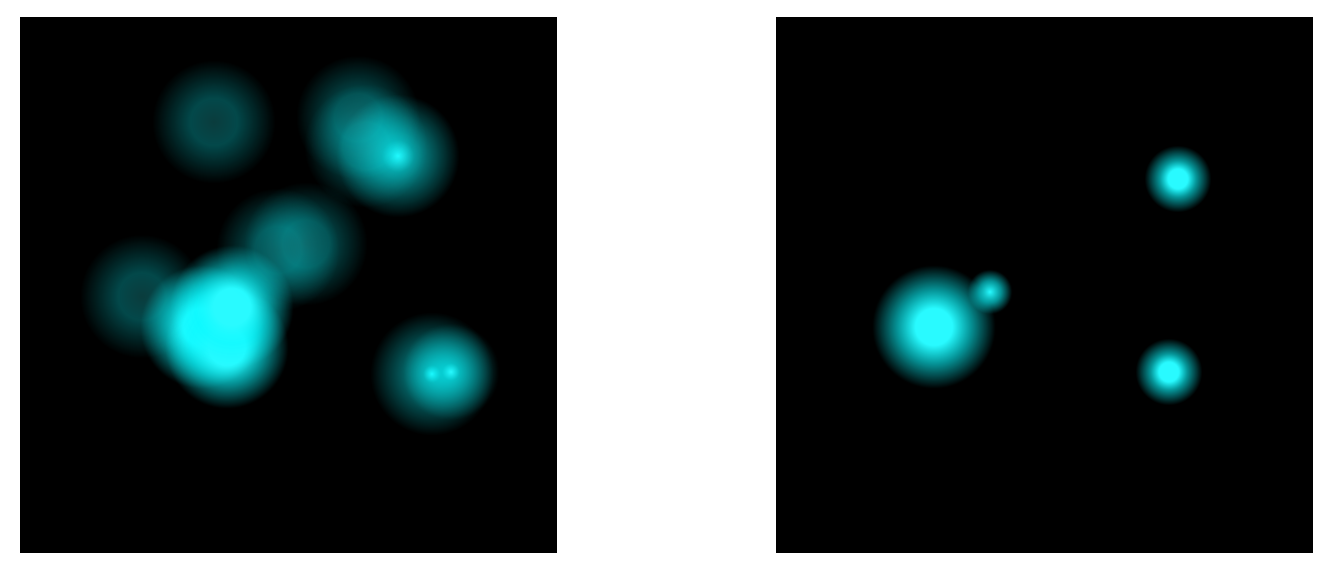}
\caption{An heat map prediction sample before the SGR filter on the left, and after the SGR filter on the right.}
\label{fig2}
\end{figure}

\textit{\textbf{Selective Gaussian Reconstruction (SGR)}}

To tackle the problems related to changing object confidences during feedback loop, we design the SGR filter. Unlike simple thresholding methods used by \cite{fairmot,1} on one channel feedback, we use a more complex filtering process with multi-channel feedback. Hence, class information is also provided during feedback by inputting the previous heat map in its full dimensions ($C \times W \times  H$). For each class $c$, we use two thresholds: $\lambda_c$ to control which local maximums will reappear on the filtered heat map; and $\theta_c$ to control which maximums will be considered as true detections. Extensive simulations show that class-specific thresholding gives better results due to unique confidence distributions of true positive detections in each class. Separation of detection $(\theta_c)$ and filtering thresholds $(\lambda_c)$ also helps the network remember not only confident detections but also local peaks that may become future detections. After deciding which centers will be reconstructed on the heat map we render each Gaussian peak with  amplified confidence scores. Confidence amplification factor $(\varphi_c)$ is also class dependent. Functionality of SGR filter is shown in \ref{eq4}, where $\hat{w}_i$ is the predicted confidence score of the local maximum and SGR filtered result is represented with $w_i$ where $\theta_c >\lambda_c$: 
 
\begin{equation} \label{eq4}
\begin{split}
w_i =\begin{cases}
0 &  \lambda_c > \hat{w}_i \\
\hat{w}_i &\theta_c > \hat{w}_i \geqslant \lambda_c\\
\hat{w}_i \times \varphi_c & \hat{w}_i \geqslant \theta_c\\
\end{cases}
\end{split}
\end{equation}
A sample heat map input and output of SGR filter are shown in Fig. \ref{fig2}.

\begin{algorithm}
	\caption{Detection and Tracking process}
	\label{Main}
	\hspace*{\algorithmicindent} \textbf{Input:} Video Frames ($I^{(t)}$) \\
    \hspace*{\algorithmicindent} \textbf{Output:} Tracklets ($\hat{o}^{(t)}$)
	\begin{algorithmic}[1]
		\For{$ t\gets 1  \ldots    $ Length of $I$ }
			\State $\hat{M},\hat{V},\hat{S} \gets $ HM-Net$(I^{(t)},I^{(t-1)},\hat{M}^{(t-1)})$
			\State $\hat{M} \gets $ NMS$ (\hat{M})$
		    \State $\mathbf{\hat{p}},\mathbf{\hat{c}}\gets$ Locate local maxima (where $\hat{M}_c >\lambda_c$ ) 
			\State $\mathbf{\hat{w}}\gets\hat{M}[\mathbf{\hat{c}},\mathbf{\hat{p}}]$, $\mathbf{\hat{v}}\gets\hat{V}[:,\mathbf{\hat{p}}]$, $\mathbf{\hat{s}}\gets\hat{S}[:,\mathbf{\hat{p}}]$
			\State  $\mathbf{\hat{m}}\gets\mathbf{\hat{p}}+\mathbf{\hat{s}}$
			\State $\hat{o}^{(t)}\gets$TRACKER$((\mathbf{\hat{m}},\mathbf{\hat{c}},\mathbf{\hat{w}},\mathbf{\hat{v}}),\hat{o}^{(t-1)},\mathbf{\theta})$
			\State $\hat{M}^{(t)}\gets$ SGR$(\mathbf{\hat{p}},\mathbf{\hat{c}},\mathbf{\lambda},\mathbf{\theta},\mathbf{\varphi})$
		\EndFor
	\end{algorithmic}
\end{algorithm}
 
Algorithm \ref{Main} summarizes the major steps of the detection and tracking process.
At each time step HM-Net uses current ($I^{(t)}$) and  previous ($I^{(t-1)}$) frames alongside filtered previous center heat map ($\hat{M}^{(t-1)}$) to predict $\hat{M},\hat{V}$, and $\hat{S}$. Then non-maxima suppression (NMS) is applied to $\hat{M}$ to extract local maxima. Local maxima are thresholded with $\lambda_c$ to determine detections on  $\hat{M}$ together with their class ($ \mathbf{ \hat{c}}$) and  position ($ \mathbf{\hat{p}}$) information. By using $\mathbf{\hat{c}}$ and $\mathbf{\hat{p}}$, confidence scores, motion and precision vectors are extracted from predicted heat maps. Precision vectors are used to update the center position coordinates of detections. Extracted information and detection threshold are sent to Tracker to get tracklets. At the end of each time step SGR function is used to filter and render the feedback heat map.

\begin{figure*}[t!]
\centerline{
\includegraphics[width=\maxwidth]{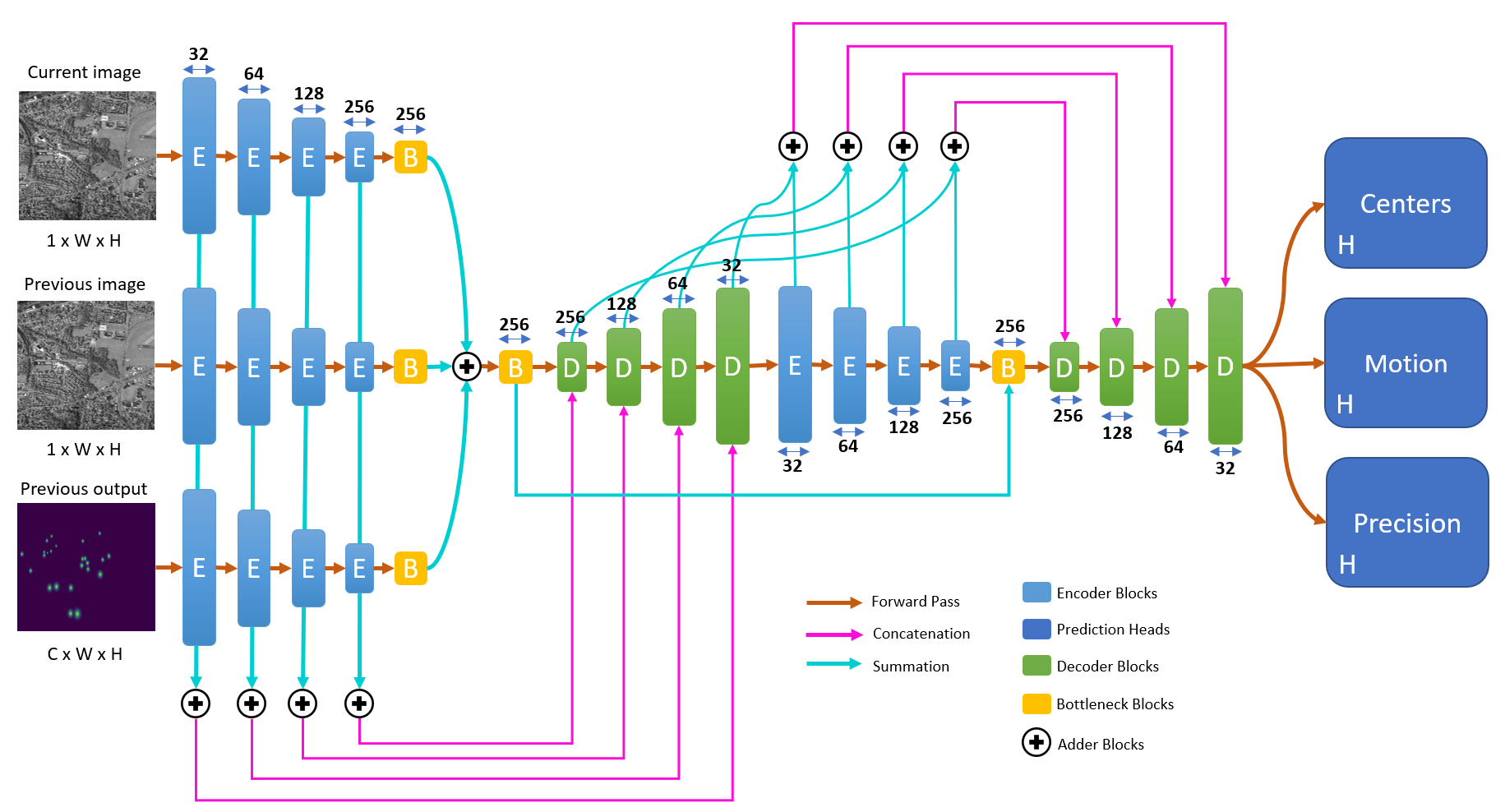}
}
\caption{The HM-Net architecture diagram with block legend.}
\label{fig3}
\end{figure*}

\subsection{HM-Net Architecture}
Center-based detection and tracking necessitates the prediction of large heat maps that contain location and class information of centers. The heat map prediction is a regression task that can be handled using deep-learning models. Overall heat map prediction task is similar to segmentation and human pose estimation tasks. These tasks can be handled using fully convolutional networks that produce output maps with dimensions proportional to input images. We have studied and tested several deep learning-based auto-encoder networks that gave state-of-the-art results on different tasks during development process of our own model, such as UNet\cite{45} and  Stacked Hourglass \cite{47} architectures, in order to combine their strengths. 

Although simple feed-forward deeper neural networks are expected to give better results, real-life tests show that deeper plain neural networks face vanishing gradient problem that hinders shallow layers' learning capability \cite{46}. Shortcuts between layers enable a more effective learning process for deeper and connected neural networks to reach expected high accuracy scores \cite{46}.  All examined architectures introduce shortcuts between layers alongside feed-forward path to increase connectivity. UNet architecture utilizes activation transfer between encoder and decoder to increase connectivity between distant layers. UNet uses concatenation for activation transfer.  Stacked Hourglass \cite{47} architecture uses summation operations between activations to ensure connectivity. Summation operation is faster; however it squeezes all activations into one. On the other hand, concatenation process yields larger activations that increase the volume of the network while preserving the information in separate channels. In HM-Net architecture, we aim to fuse these connection options for optimal trade-off between accuracy and model complexity.

\begin{figure*}[htb]
\includegraphics[width=\maxwidth]{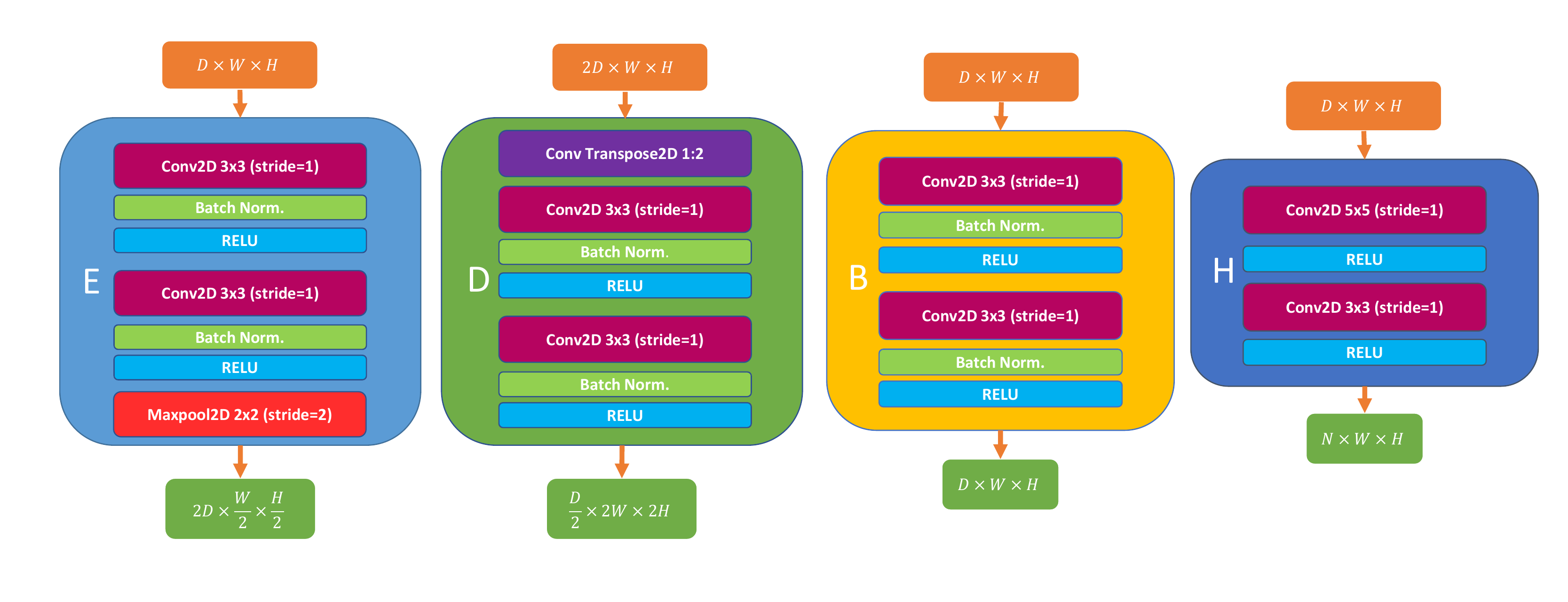}\caption{
Detailed representations of the model. The Encoder block (E),the Decoder block (D),the Bottleneck block (B) and the Head block (H) are shown with their input (orange) and output (green) dimensions. Dimensions $D\times W\times H$ stand for depth, width and height. $N$ is configurable parameter for head output depth; for Precision and Motion heads $N=2$ and for Centers Head $N$ equals to number of classes. Input depth $2D$ on encoder block is a result of concatenation based connections.\label{fig4}}
\end{figure*}

Detailed structural diagram of the HM-Net is shown on Fig. \ref{fig3}. HM-Net architecture consists of two cascaded auto-encoder sections which have several shortcuts between modules. Model is designed as one-stage tracker and detector. It can obtain matched predictions in one feed-forward pass. One-stage nature of the architecture increases overall throughput speed. Cascaded auto-encoders are used to increase the depth of the model by facilitating deeper bottleneck blocks while not using too much memory. Deeper layers enable HM-Net to extract more complex features.

HM-Net takes two consecutive frames of input video and SGR filtered version of previous heat map to exploit temporal changes. Each input has a dedicated encoder for feature extraction process. Each encoder has 4 consecutive encoder blocks (Fig.\ref{fig4}) and is followed by a deeper bottleneck layer. The role of encoder blocks is to increase the depth of activations ($\times2$) while reducing channel dimensions to decrease memory utilization. At the end of first encoder stage separately extracted features are sent to 3 parallel bottleneck blocks which are later fused into one bottleneck by using summation operation, and fed through first stage decoder. First stage decoder consists of 4 consecutive decoder blocks (Fig.\ref{fig4}). Second encoding and decoding modules have same data flow pattern with their counterparts in first auto-encoder. As Fig. \ref{fig3} shows there are several shortcuts and activation adder blocks that are introduced to increase connectivity. Activation adder blocks simply calculate summations of connected blocks' outputs. The blue arrows in first stage encoder represent summation of corresponding activations coming from three separate encoder blocks (see Fig. \ref{fig3}). Except for the first stage bottleneck activation adder block, each adder block is followed by a concatenation operation which concatenates the input activation of a decoder block and the output of connected summation block. HM-Net has three separate prediction heads to host predictions separately, which are object center detection heat map head, precision head, and motion vector head. All outputs have the same width and height with the input image.  

\subsection{HM-NET Training}
WPAFB 2009 \cite{36} dataset is used for both model training and evaluation. WPAFB  dataset consists of roughly $25,000\times20,000$ grayscale stitched images with NITF format. In order to reduce computational complexity, the given set of images are downsampled by a factor of 2 to the average size of $12,500\times10,000$ pixels. We apply several preprocessing steps to obtain training-ready images. Images are converted to PNG format,and then a simple registration algorithm is applied to compensate for the camera motion. The training set consists of cropped versions images which have $544\times960$ dimensions. Smaller cropped training images reduces computational complexity and memory footprint during training. This enables usage of larger batch sizes during training.    

As stated in Section \ref{sec:RDTa}, our model uses previous image and output as feedback to exploit temporal and spatial information together. During inference,  the SGR filter is applied on predicted heat map to improve prediction performance.
SGR filter yields a reconstructed feedback heat map consisting of 2D Gaussian peaks according to their predicted confidence values. During training, we have to simulate the actual feedback heat map of the model inference. This simulation should be realistic enough to encourage the model to learn corner cases that cause false positive propagation, vanishing confidence and location precision loss. 

False positive propagation process is shown with three heat maps in Fig. \ref{figFPP}. At first glance distribution of predicted confidence values have high variance and they are hard to simulate realistically during training. For feedback simulation in training, we use ground truth object centers of previous image when constructing the heat map. Setting confidence scores of ground truth objects as 1 does not lead to an effective training. In our initial trials, confidence scores $\hat{w}_i$ of fed-back detections were observed to be varying around $0.6$. Since predictions during inference have varying confidence scores, a randomized confidence generation method that can simulate observed score variations is implemented for better feedback simulation during model training. 

SGR filter ensures that heat map outputs have varying confidence scores but solid 2D Gaussian shapes as discussed above. Solidified Gaussian curves are more imitable than raw heat map outputs (see Fig.\ref{fig2}). Our primary aim on training is simulating SGR filtered feedback heat maps in a realistic setup with several advanced data augmentations. These augmentation methods, such as random location shift and random center deletion, are key to increase generalization capability and emulate model inference. Random confidence reduction (RCR) method is applied on center confidences to simulate more realistic confidence scores on a feedback scenario.  RCR applies up to $0.8$ confidence reduction with $70\%$ probability to each confidence ground truth $w_i$. The reduction factor and probability are set based on ablation studies to maximize training performance.    

False positive propagation problem is shown in Fig. \ref{figFPP}. Around one detection we see tree root shaped spreading false detection clusters. To reduce the occurrence of this unwanted situation we implement randomized center propagation (RCP). During heat map simulation RCP picks random labels and places several false detections around the ground truth to simulate the propagation problem. RCP is also used to place false positive object centers randomly on the feedback heat map to teach our network how to eliminate all types of false positive detections.

Data distributions among heat maps are imbalanced; in other words Gaussian peaks cover very small portion of heat maps. That is why we utilize a modified focal loss function from \cite{5,1,42,43}. The loss $L_H$ is computed between the ground truth heat map $M_{c,x,y}$ and its prediction $\hat{M}_{c,x,y}$ for each class and at every pixel location:

\begin{equation} \label{eq5}
L_H = -\frac{1}{N} \sum_{c,x,y} f(M_{c,x,y}, \hat{M}_{c,x,y}) 
\end{equation}
where
\begin{equation}\label{eq6}
  f(m,n) \! = \! \left\{
    \begin{array}{lr}
       (1-n)^\alpha   \log (n), & \mbox{if } m=1   \\
        (1-m)^\beta (n)^\alpha \log (1-n), & \mbox{otherwise}
    \end{array}
    \right.
\end{equation}
where $N$ is the total count of labeled objects and hyper-parameters of focal loss are set to $\alpha=2$, $\beta=4$.

For learning motion vectors and subpixel center locations, we use supervised linear loss ($L_1$ loss):
\begin{equation} \label{eq7}
\begin{split}
L_{\mathbf{x}}=\left(\frac{1}{N}\right)\sum_{i=1}^N|\hat{\mathbf{x}}_{i}-\mathbf{x}_{i}\lvert 
\end{split}
\end{equation}
where $\mathbf{x}_i$ is $\mathbf{v}_i$ for motion vectors and $\mathbf{x}_i$ is $\mathbf{s}_i$ for subpixel locations. The weighted sum of those loss functions is used during training. Weights are set by considering the importance of the corresponding head while maintaining training balance. For model optimization  Adam optimizer is used with a learning rate of $0.0013$. Model training with batch size 6 is concluded at 16\textsuperscript{th} epoch to prevent overfitting.

\begin{figure}[htpb]
\includegraphics[width=\maxwidth]{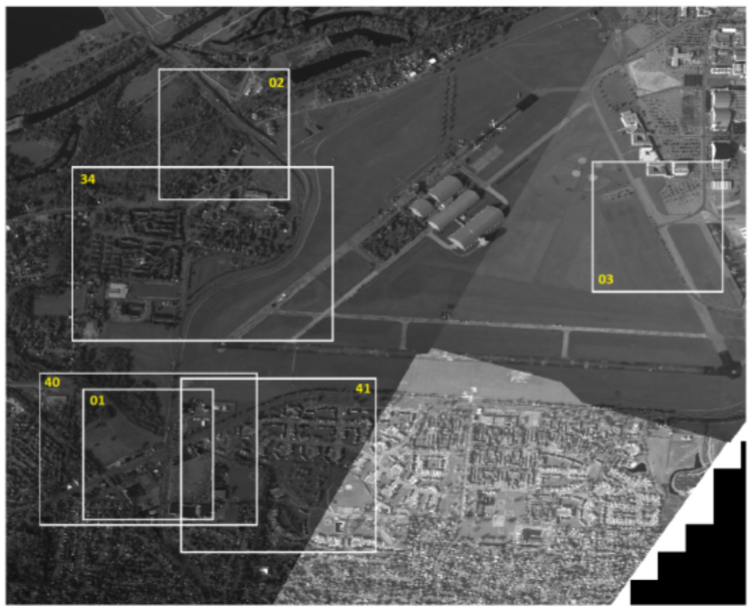}
\caption{Locations of cropped AOI's in a complete WAMI frame. Courtesy of \cite{19}.}
\label{fig5}
\end{figure}

\begin{figure*}[htpb]
\includegraphics[width=\maxwidth]{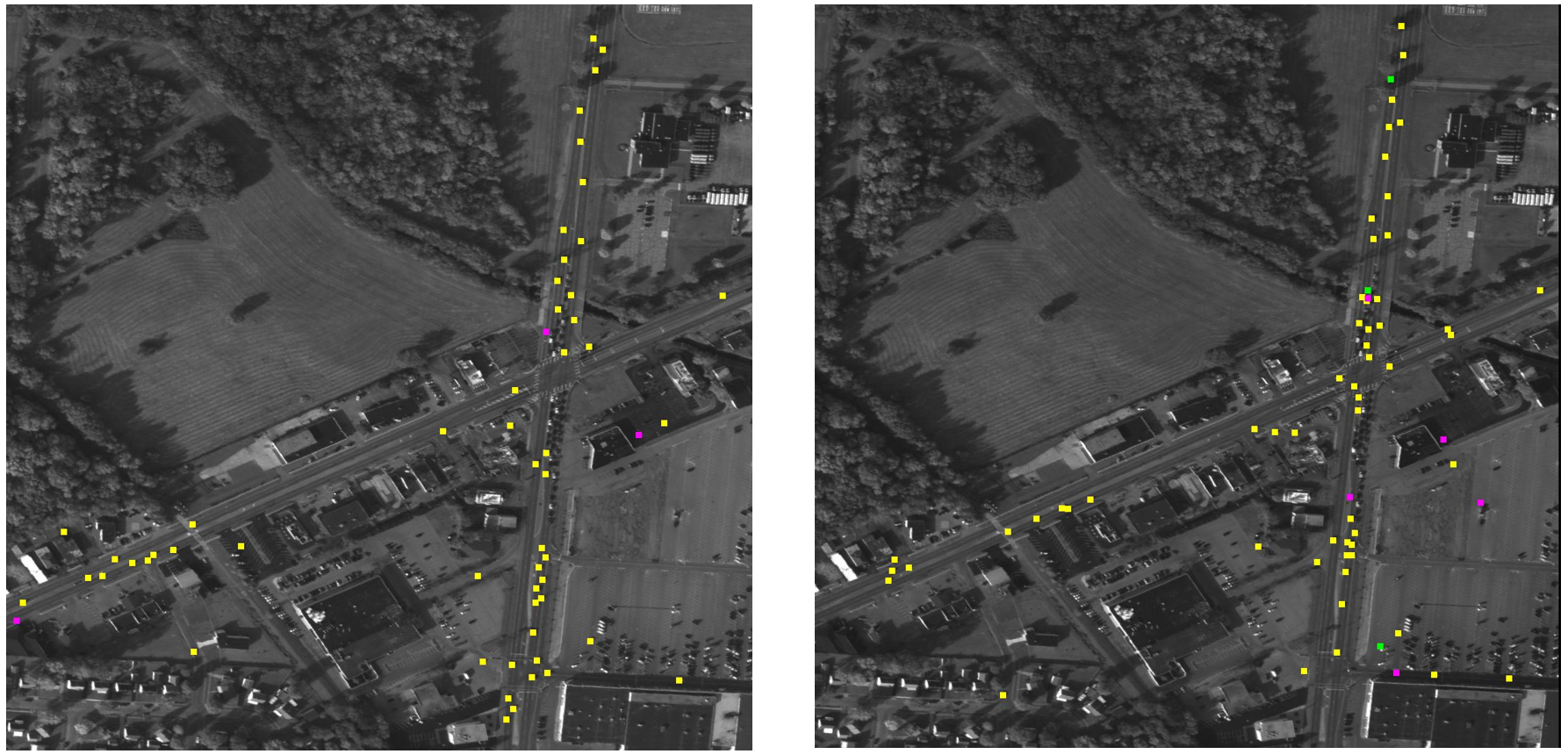}
\caption{Visualization of true positives (yellow), false positives (green), and false negatives (magenta). High-resolution versions can be viewed at \url{https://sens.medipol.edu.tr/visionai-was/hm-net-detection-visual-results/}}
\label{figxi}
\end{figure*}

\section{Results and Discussions}
\label{sec:RAD}
In this section, first we discuss our detection and tracking results, in comparison with state-of-the-art methods from literature. Then an ablation study is presented to illustrate the effects of hyper-parameters and novel data augmentation methods on the  detection and tracking performances.

\subsection{Evaluation of Detection and Tracking Perfomance}
Our tests are conducted on the WPAFB dataset \cite{36} at predetermined cropped AOIs (see Fig. \ref{fig5} and Table \ref{table_crops}). A detection is considered as true positive when there is at least one ground truth point within 10 pixels range and each ground-truth point can only be assigned to one detection (see \cite{24}). Table \ref{tab:table2} provides F1 scores for HM-Net and other state-of-the-art object detectors from literature in 6 different tested AOIs. HM-Net provides the highest F1 score in 5 regions and comes in second place for AOI 40. In Fig. \ref{figxi} locations of true positive, false positive and false negative detections are shown on two sample frames.  
As discussed in Section \ref{sec:RDT}, HM-Net is trained to detect both moving and stationary vehicles, which helps in reducing false moving object detections. Stationary and moving vehicle detections are visualized in Fig.\ref{figxx}.

Table \ref{tab:table03} provides precision and recall values for HM-Net, \cite{24} and \cite{1}. (For training CenterTrack \cite{1} in WPAFB dataset, bounding box prediction head of the architecture is ignored.) HM-Net is superior to \cite{24} in all tested regions. Precision values for  \cite{1} and HM-Net are similar, but HM-Net is superior to  \cite{1} in terms of recall.

\begin{table}[ht]
\centering
\caption{Size and location information of AOIs.}
\label{table_crops}
\begin{tabular}{|c|c|c|c|}
\hline
\textbf{AOI}                       & \textbf{Center} & \textbf{Width} & \textbf{Height} \\ \hline
01 &(4700, 8050)& 1200& 1200  \\
02 &(5650, 5300)& 1200 &1200   \\
03 &(9400, 5950)& 1200& 1200         \\
34 &(5300, 6300)& 2400& 1600        \\
40 &(4850, 8050)& 2000& 1400          \\
41 &(6000, 8250)& 1800& 1600          \\ \hline
\end{tabular}
\end{table}

Table \ref{tab:table4} compares the same three methods in terms mean Average Precision (mAP), which corresponds to the area under precision-recall curve as defined by object detection evaluation procedures \cite{visD}. mAP is calculated for two scenarios, by requiring either a maximum of 10 pixels ($d<10$) or 5 pixels ($d<5$) distance between matching detected and ground truth centers. Clearly, HM-Net surpasses both \cite{24} and \cite{1} in all AOIs and the performance difference is even more dramatic when $d<5$ is required. \cite{24} fails to provide precise center coordinates for most of the detected objects. While both HM-Net and CenterTrack \cite{1} produce accurate object centers, HM-Net produces more precise  centers due to the use of an extra Precision Head at the network output. The average distance of detected centers from ground truth coordinates is 1.8 pixels. Even though we upscale predictions at the network output, we surpass \cite{24} and \cite{25}, which have average distances of 3.8 pixels and roughly 2 pixels, respectively.

\begin{table}[htp]
\centering
\caption{The detection performance (F1 scores) comparison table for six AOIs of WPAFB dataset \cite{38}. }
\begin{tabular}{|c|cccccc|}
  \hline
\textbf{Method} & \textbf{01} & \textbf{02} & \textbf{03} & \textbf{34} & \textbf{40} & \textbf{41} \\ \hline
\cite{saleemi} & 0.783	&0.793 &	0.876 &	0.755 &	0.749 &	0.762\\ 
\cite{xiao} & 0.738 & 0.820  & 0.868  &  0.761  & 0.733 & 0.700\\ 
\cite{34} &0.866& 0.890 & 0.900 & - & - & -\\ 
\cite{32} &0.850 & 0.876 & 0.889   & 0.826  &0.817 & 0.799\\ 
\cite{keck} & 0.743 & 0.825& 0.876 & 0.763 & 0.737 & 0.708 \\
\cite{30} & -& -& -&  0.900&	0.900&	0.930\\
\cite{25} &0.947& 0.951& 	0.942& 	0.933&	\textbf{0.983} &	0.928\\
\cite{24}& 0.935& 0.947& 0.945 &0.953 &0.935& 0.934\\
\cite{1}&0.944 &0.967 &0.964 &0.967& 0.938& 0.948\\\hline
\textbf{HM-Net} & \textbf{0.952} & \textbf{0.971} & \textbf{0.972} & \textbf{0.973} & 0.953 & \textbf{0.950}\\ \hline
\end{tabular}
\label{tab:table2}
\end{table}

 
\begin{table}[ht]
\centering
\caption{Comparison of average precision and recall values of \cite{24} and \cite{1} with our results for the six AOIs.}

\begin{tabular}{|c|ccc|ccc|}
\hline
\multirow{2}{*}{\textbf{AOI}} & \multicolumn{3}{c|}{\textbf{Precision}} & \multicolumn{3}{c|}{\textbf{Recall}} \\ \cline{2-7} 
  & \cite{24}  & \cite{1}  & HM-Net  & \cite{24} & \cite{1} & HM-Net \\ \hline
01& 0.939 & \textbf{0.979} & 0.977 & 0.914 & 0.896 & \textbf{0.929}\\
02& 0.958 & 0.981 & \textbf{0.985} & 0.932 & 0.950 &\textbf{0.959}\\
03& 0.933 & 0.981 & \textbf{0.987} & 0.951 & 0.945 & \textbf{0.958}\\
34& 0.956 & 0.985 & \textbf{0.993} & 0.941 & 0.944 & \textbf{0.955}\\
40& 0.943 & \textbf{0.982} & 0.980 & 0.911 & 0.877 & \textbf{0.926}\\
41& 0.945 & 0.983 & \textbf{0.984} & 0.919 & 0.911 & \textbf{0.920} \\ \hline
\end{tabular}
\label{tab:table03}
\end{table}

 
\begin{table}[ht]
\centering
\caption{Comparison of mAPs (in percentage) of \cite{24} and \cite{1} with our results for the six AOIs.}
\label{tab:my-table}
\begin{tabular}{|c|ccc|ccc|}
\hline
\multirow{2}{*}{\textbf{AOI}} & \multicolumn{3}{c|}{\textbf{mAP @$d<10$}} & \multicolumn{3}{c|}{\textbf{mAP @$d<5$}} \\ \cline{2-7} 
  & \cite{24}  & \cite{1}  & HM-Net  & \cite{24} & \cite{1} & HM-Net \\ \hline
01& 86.41 & 87.72 & \textbf{92.71} & 54.20 & 85.34 & \textbf{91.12} \\
02& 90.16 & 93.66 & \textbf{95.91} & 57.82 & 92.22 & \textbf{95.15} \\
03& 89.22 & 93.05 & \textbf{96.09} & 63.72 & 92.51 & \textbf{95.59} \\
34& 90.59 & 93.37 & \textbf{95.90} & 59.92 & 90.64 & \textbf{94.38} \\
40& 86.30 & 86.26 & \textbf{92.41} & 52.55 & 83.94 & \textbf{90.80} \\
41& 87.35 & 89.75 & \textbf{93.27} & 50.08 & 86.54 & \textbf{91.84} \\ \hline
\end{tabular}
\label{tab:table4}
\end{table}

\begin{table}[ht]

\caption{Comparison of tracking performance for the six AOIs with tracking mAP from \cite{visD}.}

\label{track_tab}
\begin{adjustbox}{width=8.6cm}

\begin{tabular}{|c|c|cccccc|}
\hline
                                              \textbf{Metrics}  & \textbf{Methods} &\textbf{01}&\textbf{02}&\textbf{03}&\textbf{34}&\textbf{40}&\textbf{41}\\ \hline
\multirow{2}{*}{\vtop{\hbox{\strut mAP}\hbox{\strut @$0.25$}}} &\cite{1}&  \textbf{78.71} & \textbf{92.30} & \textbf{82.66} & 87.92 & 72.97 & \textbf{77.64}\\
\multicolumn{1}{|c|}{}                              &HM-Net &      78.39 & 90.31 & 80.74 & \textbf{90.86} & \textbf{79.10} & 77.21 \\ \hline
\multirow{2}{*}{\vtop{\hbox{\strut mAP}\hbox{\strut @$0.50$}}}                       &\cite{1} & 33.44 & 67.22 & 63.04 &68.10 & 26.26 & 53.03 \\
                                                    &HM-Net &      \textbf{47.04}   & \textbf{83.31}   & \textbf{69.77}   & \textbf{82.10}   & \textbf{47.54}  &  \textbf{70.38}  \\ \hline
\multirow{2}{*}{\vtop{\hbox{\strut mAP}\hbox{\strut @$0.75$}}}                       &\cite{1} & 10.57   & 42.54    & 18.19   & 40.56   &  6.05   &  20.84  \\
                                                    &HM-Net     &  \textbf{16.73}    & \textbf{51.96}   & \textbf{31.90}   & \textbf{51.13}   & \textbf{19.89} & \textbf{42.89} \\ \hline
\multirow{2}{*}{mAP}                                &\cite{1} & 40.90 & 67.35 & 54.63 &65.53 &35.10 &50.50\\ 
                                                    &HM-Net&       \textbf{47.39}& \textbf{75.19} & \textbf{60.80} & \textbf{74.70} &\textbf{48.84}&\textbf{63.49}\\ \hline 
\end{tabular}

\end{adjustbox}
\end{table} 

Table \ref{track_tab} compares the tracking accuracy of HM-Net against CenterTrack (other methods could not be tested because their tracking codes were not available).  mAP for tracking is evaluated as described in \cite{visD}. Tracks are sorted based on their length, i.e. longer tracks are assigned higher confidence and matched first. HM-Net provides significantly better track matching accuracy than \cite{1} in all AOIs. Simulation results also show that HM-Net tracking IoU (Intersection over Union) is significantly better than that of \cite{1}. This means that HM-Net tracks provide on average a longer overlap with the ground truth tracks. This could also be seen by comparing mAP results of both methods at 0.50 and 0.75 IoU, for which  HM-Net is far superior.

\begin{figure}[ht]\includegraphics[width=\maxwidth]{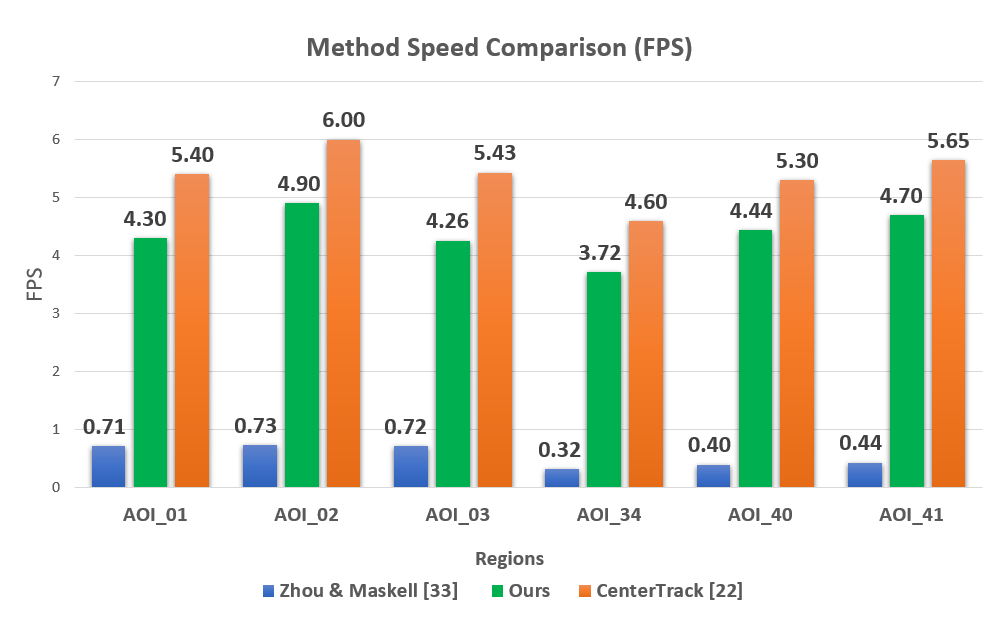}
\caption{Speed (fps) comparison of different methods on AOIs.\label{fig6}}
\end{figure}

Fig. \ref{fig6} provides inference speeds for HM-Net, \cite{24} and \cite{1} in terms of fps  for different AOIs on a RTX2080Ti 11GB GPU. HM-Net out-performs \cite{24} in speed by roughly 7 times. Although HM-Net is about  30\% slower than \cite{1},  better detection and tracking results justify this slightly higher execution times. Since both HM-Net and \cite{1} algorithms are faster than typical WAMI sensor frame rates, both models could perform real-time object detection in a medium resolution WAMI sequence. 

In \cite{24}, the model utilizes background subtraction followed by CNN-based false positive filtering from cropped detection regions. As a result the model speed in \cite{24} is vulnerable to image size and total target count variations. However, image size variations and tracking target count do not much affect the inference speed of HM-Net, due to its one-stage network, which processes whole inputs in a single feed-forward pass.

\subsection{Ablation Study}
In this section we investigate the effect of several hyper-parameters on the model's performance for object detection and tracking. In our experiments we have tested  various combination of these parameters to see their effects on overall model accuracy. Selected experiment samples will be discussed below to evaluate the effects of detection threshold ($\theta_c$), combination of confidence boosting factor ($\varphi_c$) and feedback thresholding ($\lambda_c$), non maximum suppression (NMS) window size, and RCP/RCR during training.

\begin{figure}
\includegraphics[width=\maxwidth]{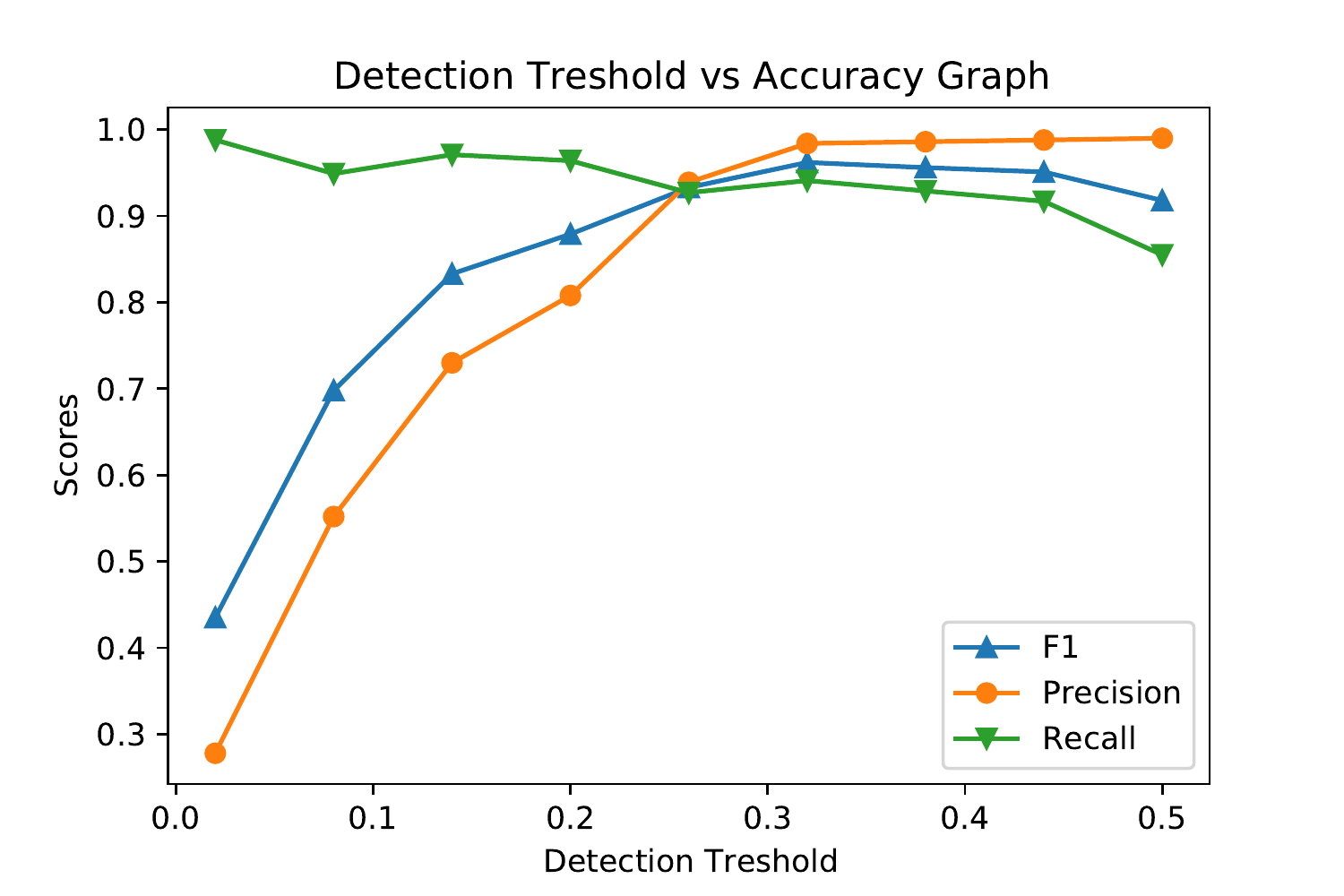}
\caption{Effect of moving vehicle detection threshold ($\theta_c$) on model accuracy. \label{figDetG}}
\end{figure}
Figure \ref{figDetG} shows the detection performance with respect to the detection threshold ($\theta_c$). As seen from the figure, detection threshold ($\theta_c$) and recall scores of the model are inversely correlated. Lower $\theta_c$ facilitates overall higher number of detections, resulting in an increase in the number of true positive detections at cost of more false positive detections, thereby decreasing F1 and precision scores. When $\theta_c$ is between $0.25$ and $0.42$ we observe a high accuracy plateau in the figure, which shows that the model performance is robust against changing detection threshold in this range. We have used $\theta_c=0.32$ in all other experiments, since it is relatively at the center of the plateau and also yields better scores on validation.

\begin{table}[ht]
\centering
\caption{Effect of feedback filtering threshold ($\lambda_c$) and confidence  amplification  factor ($\varphi_c$) values on detection accuracy.}
\label{tablebf}

\begin{tabular}{|c|c|ccc|}
\hline
$\lambda_c$           & $\varphi_c$ & \multicolumn{1}{c}{\textbf{F1}} & \multicolumn{1}{c}{\textbf{Precision}} & \textbf{Recall} \\ \hline
\multirow{2}{*}{32} & 1.0  & 0.950 & 0.986  & 0.919  \\ \cline{2-2}
                                    & 1.2   & 0.953 & \textbf{0.998}   & 0.912  \\ \cline{1-2}
\multirow{2}{*}{28} & 1.0 & 0.957 & 0.974 & 0.939  \\ \cline{2-2}
                                    & 1.2 & \textbf{0.962} & 0.984  & \textbf{0.941}  \\  \cline{1-2}
\multirow{2}{*}{24} & 1.0     & 0.939  & 0.945   & 0.933  \\ \cline{2-2}
                                    & 1.2  & 0.948 & 0.972  & 0.925  \\\hline
\end{tabular}
\end{table}
On Table \ref{tablebf} we can observe the effect of different $\lambda_c$ and $\varphi_c$  combinations on F1, recall, and precision detection scores, where all other parameters are fixed to their best known values. The use of separate feedback filter and detection thresholds is proposed to be able to represent objects below the detection threshold on feedback heat maps (see Section \ref{sec:RDT}). The motivation in keeping those objects on the heat map is to encourage their reappearance as valid object detections in upcoming frames. Note that $\varphi_c=1$ and $\lambda_c=0.32$ on Table \ref{tablebf} corresponds to the base case where there is no confidence boosting and no SGR filtering, i.e. $\lambda_c=\theta_c$. Lowering $\lambda_c$ with respect to $\theta_c$ increases recall and reduces precision, yet resulting in a higher overall F1 score. However, as $\lambda_c$ is further reduced, the performance starts to deteriorate. An optimal value of $\lambda_c=0.28$ puts on the feedback heatmap additional detections that have the potential of reappearing in upcoming frames, and yet at same time does not increase the number of false positive detections overwhelmingly. Amplifying confidences of detected objects on their feedback representations increases model precision and F1 scores as well. The best combination of parameters is at $\lambda_c=0.28$ and $\varphi=1.2$, which corresponds to a 1.2\% improvement in F1 score over the base case.     

\begin{figure}[ht]
\includegraphics[width=\maxwidth]{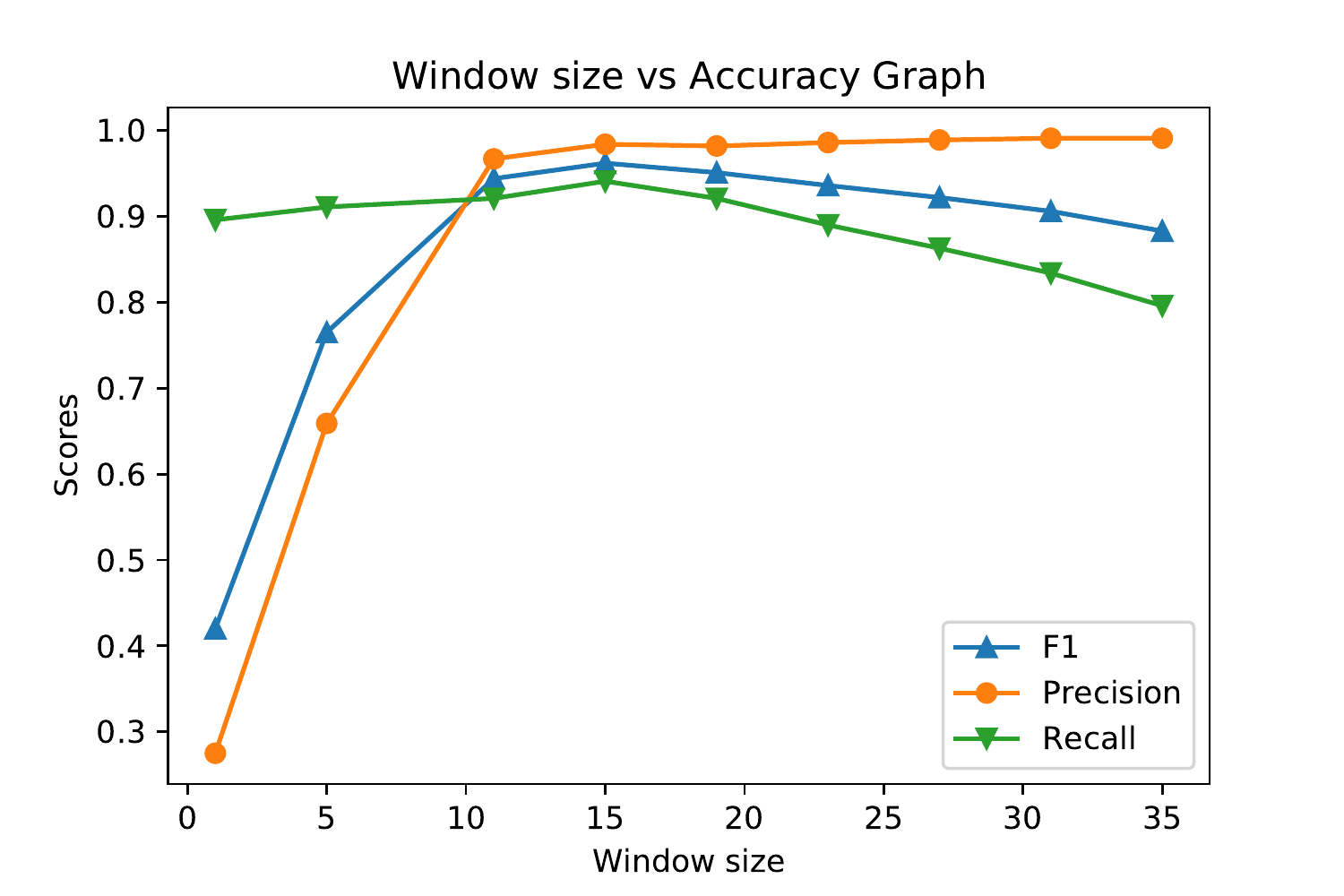}
\caption{Effect of NMS window size on model accuracy.\label{figWinG}}
\end{figure}

As discussed in Section \ref{sec:RDTa}, NMS window size is crucial for object detection and tracking operations. The primary role of the window is to eliminate adjacent high confidence scores close to the local maxima. However choosing the  optimal window is not a straightforward task. As Figure \ref{fig2} shows the model does not yield perfect Gaussian shapes on the center location predictions. A large window size ensures the elimination of adjacent high confidence values that are on the same object, but might also eliminate the centers of other close by objects. On the other hand, for small window sizes, nearby high confidence pixels that belong to the same object can be misinterpreted as centers of separate objects. Fig. \ref{figWinG} justifies these discussions. For smaller window sizes we observe high number of false positives and a relatively high recall score. As window size exceeds $11\times 11$, the model enters a steady high accuracy plateau. However F1 score starts to descend as window size exceeds $15\times 15$. This negative trend for larger window sizes is a result of a slowing decrease in false positives and fast increase in false negatives. As a result, $15\times 15$ is selected as NMS window size during both detection and tracking operations. Note that this optimal window size is in line with the 10-15 pixel vehicle length in WPAFB dataset.
\begin{table}[htb]
\centering
\caption{ Effect of RCP/RCR training methods on detection and tracking accuracy.}

\begin{tabular}{|c|cc|cc|}
\hline
\multirow{2}{*}{\textbf{AOI}} & \multicolumn{2}{c|}{\textbf{mAP$(\%)$ @$d<10$}}&\multicolumn{2}{c|}{\textbf{mAP$(\%)$}(Tracking)} \\ \cline{2-5} 
  & Base Model  & HM-Net  &Base Model& HM-Net \\ \hline 
01&  89.93 & \textbf{92.71} & 44.81 & \textbf{47.39}\\
02&  95.33 & \textbf{95.91} & 74.67 & \textbf{75.19}\\
03&  95.35 & \textbf{96.09} & 60.55 & \textbf{60.80}\\
34&  95.41 & \textbf{95.90} & 73.91 & \textbf{74.70}\\
40&  87.98 & \textbf{92.41} & 45.74 & \textbf{48.84}\\
41&  91.94 & \textbf{93.27} & 61.33 & \textbf{63.49}\\ \hline

\end{tabular}
\label{tab:table05}
\end{table}

Table \ref{tab:table05} investigates the performance contributions of RCP and RCR methods applied during training of the model. Introduction of  RCP and RCR aims to reduce false positive detection count and give model a better simulation of the feedback process during training (see Section \ref{sec:RDT}). Effect of RCP and RCR is tested by comparing against a Base Model  that is completely identical but trained without RCP and RCR data augmentations. Trained base model is tested under identical testing conditions. Results are shown on Table \ref{tab:table05}. 
Introduction of RCP and RCR increases model success in both detection and tracking. We observe significant improvement in scores, especially for AOI-01 and AOI-40. These AOIs have signalized busy intersections, which causes vehicles to stop and move suddenly. Also intersections can cause problems such as  tracking id switches due to sharp changes in vehicle headings. With the help of RCP and RCR in training, the model learns to handle sudden transitions between motion states of the vehicles and alterations in their headings better than plainly trained version. 

\begin{figure*}[t]
\includegraphics[width=\maxwidth]{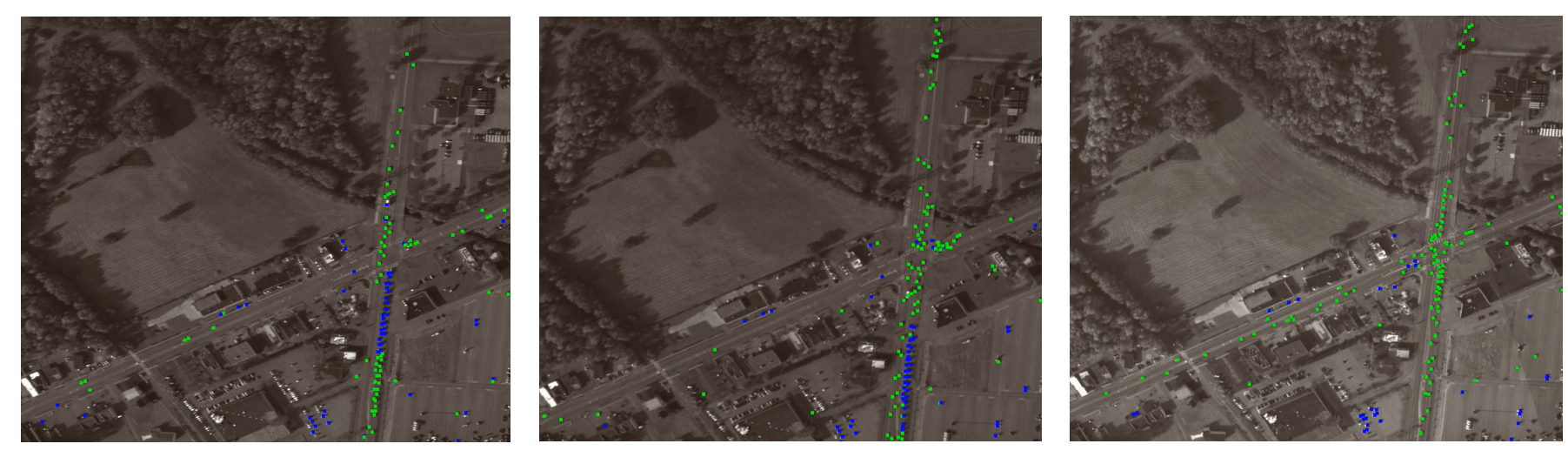}
\caption{Visualization of moving (green) and stationary (blue) objects. Behavior of the vehicles can be observed on a signalized junction.   \label{figxx}}
\end{figure*}

\color{black}
\section{Conclusion and Future Work}
\label{sec:CFW}

In this work, we have proposed a combined detection and tracking method that utilizes center-based object detection and tracking. Our method relies on our one-stage CNN based architecture called HM-Net, which generates heat maps of object center locations and object displacements for detection and tracking tasks. Heat map-based approach facilitates simultaneous detection of high and varying number of objects with fast and stable speed. Our method uses both current and previous frames and filtered previous predictions as input to exploit spatio-temporal information. Novel model training tools and feedback filters address existing problems on the training and inference of heat map-based detectors, and these methods increase overall model performance, while reducing the number of false positive detections. 

Our approach is significantly faster than frame differencing and background subtraction-based state-of-the-art WAMI detectors and yields compatible speed with one-stage counterparts. HM-Net yields state-of-the-art results on simulations conducted on WPAFB dataset \cite{36} for detection and tracking of moving vehicles. To improve the overall performance, alterations in model architecture and their implications on both tracking and detection performances will be further examined.Currently HM-Net uses just a single prior frame and single prior heatmap; utilizing a higher number of prior frames and outputs will be investigated to increase overall performance. A complex vehicle motion model, in the form of a Kalman filter, could also be integrated to improve object tracking accuracy. Model will also be improved to work with larger resolution images that cover larger WAMI regions. Also, MOT applications for aerial imagery at different altitudes (i.e. drone videos) of HM-Net architecture will be investigated as future work.  

\bibliographystyle{IEEEtran}
\bibliography{references}

\begin{thebibliography}{10}
\providecommand{\url}[1]{#1}
\csname url@samestyle\endcsname
\providecommand{\newblock}{\relax}
\providecommand{\bibinfo}[2]{#2}
\providecommand{\BIBentrySTDinterwordspacing}{\spaceskip=0pt\relax}
\providecommand{\BIBentryALTinterwordstretchfactor}{4}
\providecommand{\BIBentryALTinterwordspacing}{\spaceskip=\fontdimen2\font plus
\BIBentryALTinterwordstretchfactor\fontdimen3\font minus
  \fontdimen4\font\relax}
\providecommand{\BIBforeignlanguage}[2]{{%
\expandafter\ifx\csname l@#1\endcsname\relax
\typeout{** WARNING: IEEEtran.bst: No hyphenation pattern has been}%
\typeout{** loaded for the language `#1'. Using the pattern for}%
\typeout{** the default language instead.}%
\else
\language=\csname l@#1\endcsname
\fi
#2}}
\providecommand{\BIBdecl}{\relax}
\BIBdecl

\bibitem{25}
R.~LaLonde, D.~Zhang, and M.~Shah, ``{Clusternet: Detecting small objects in
  large scenes by exploiting spatio-temporal information},'' in \emph{{IEEE
  Conference on Computer Vision and Pattern Recognition (CVPR)}}, 2018, pp.
  4003--4012.

\bibitem{31}
T.~T. Nguyen, H.~Grabner, H.~Bischof, and B.~Gruber, ``{On-line boosting for
  car detection from aerial images},'' in \emph{{IEEE International Conference
  on Research, Innovation and Vision for the Future (RIVF)}}, 2007, pp. 87--95.

\bibitem{33}
R.~Ruskon{\'e}, L.~Guigues, S.~Airault, and O.~Jamet, ``{Vehicle detection on
  aerial images: A structural approach},'' in \emph{{IEEE International
  Conference on Pattern Recognition (ICPR)}}, vol.~3, 1996, pp. 900--904.

\bibitem{35}
T.~Zhao and R.~Nevatia, ``Car detection in low resolution aerial images,''
  \emph{Image and Vision Computing}, vol.~21, pp. 693--703, 2003.

\bibitem{36}
U.~S. Air Force Research~Laboratory, ``{Wright-Patterson Air Force Base (WPAFB)
  dataset},'' 2009.

\bibitem{17}
G.~Kopsiaftis and K.~Karantzalos, ``{Vehicle detection and traffic density
  monitoring from very high resolution satellite video data},'' in \emph{{IEEE
  International Geoscience and Remote Sensing Symposium}}, 2015, pp.
  1881--1884.

\bibitem{30}
\BIBentryALTinterwordspacing
R.~P. Pflugfelder, A.~Weissenfeld, and J.~Wagner, ``On learning vehicle
  detection in satellite video,'' \emph{CoRR}, vol. abs/2001.10900, 2020.
  [Online]. Available: \url{https://arxiv.org/abs/2001.10900}
\BIBentrySTDinterwordspacing

\bibitem{2}
W.~Liu, D.~Anguelov, D.~Erhan, C.~Szegedy, S.~Reed, C.-Y. Fu, and A.~C. Berg,
  ``{SSD: Single Shot MultiBox Detector},'' in \emph{European Conference on
  Computer Vision (ECCV)}.\hskip 1em plus 0.5em minus 0.4em\relax Springer,
  2016, pp. 21--37.

\bibitem{3}
K.~He, G.~Gkioxari, P.~Dollar, and R.~Girshick, ``{Mask R-CNN},'' \emph{{IEEE
  Transactions on Pattern Analysis and Machine Intelligence}}, vol.~42, no.~2,
  pp. 386--397, 2020.

\bibitem{4}
R.~Girshick, ``{Fast R-CNN},'' in \emph{{IEEE International Conference on
  Computer Vision (ICCV)}}, 2015, pp. 1440--1448.

\bibitem{5}
\BIBentryALTinterwordspacing
X.~Zhou, D.~Wang, and P.~Kr{\"{a}}henb{\"{u}}hl, ``Objects as points,''
  \emph{CoRR}, vol. abs/1904.07850, 2019. [Online]. Available:
  \url{http://arxiv.org/abs/1904.07850}
\BIBentrySTDinterwordspacing

\bibitem{6}
R.~Girshick, J.~Donahue, T.~Darrell, and J.~Malik, ``{Rich feature hierarchies
  for accurate object detection and semantic segmentation},'' in \emph{{IEEE
  Conference on Computer Vision and Pattern Recognition (CVPR)}}, 2014, pp.
  580--587.

\bibitem{7}
S.~Ren, K.~He, R.~Girshick, and J.~Sun, ``{Faster R-CNN: Towards real-time
  object detection with region proposal networks},'' \emph{{Advances in Neural
  Information Processing Systems}}, vol.~28, pp. 91--99, 2015.

\bibitem{8}
Z.~Cai and N.~Vasconcelos, ``{Cascade R-CNN: Delving into high quality object
  detection},'' in \emph{{IEEE Conference on Computer Vision and Pattern
  Recognition (CVPR)}}, 2018, pp. 6154--6162.

\bibitem{9}
J.~Redmon, S.~Divvala, R.~Girshick, and A.~Farhadi, ``{You Only Look Once:
  Unified, real-time object detection},'' in \emph{IEEE Conference on Computer
  Vision and Pattern Recognition (CVPR)}, 2016, pp. 779--788.

\bibitem{10}
\BIBentryALTinterwordspacing
C.~Fu, W.~Liu, A.~Ranga, A.~Tyagi, and A.~C. Berg, ``{DSSD} : Deconvolutional
  single shot detector,'' \emph{CoRR}, vol. abs/1701.06659, 2017. [Online].
  Available: \url{http://arxiv.org/abs/1701.06659}
\BIBentrySTDinterwordspacing

\bibitem{11}
N.~Wojke, A.~Bewley, and D.~Paulus, ``{Simple online and realtime tracking with
  a deep association metric},'' in \emph{{IEEE International Conference on
  Image Processing (ICIP)}}, 2017, pp. 3645--3649.

\bibitem{12}
X.~Hou, Y.~Wang, and L.-P. Chau, ``{Vehicle tracking using Deep Sort with low
  confidence track filtering},'' in \emph{{IEEE International Conference on
  Advanced Video and Signal Based Surveillance (AVSS)}}, 2019, pp. 1--6.

\bibitem{fairmot}
\BIBentryALTinterwordspacing
Y.~Zhang, C.~Wang, X.~Wang, W.~Zeng, and W.~Liu, ``A simple baseline for
  multi-object tracking,'' \emph{CoRR}, vol. abs/2004.01888, 2020. [Online].
  Available: \url{https://arxiv.org/abs/2004.01888}
\BIBentrySTDinterwordspacing

\bibitem{14}
K.~Kang, H.~Li, J.~Yan, X.~Zeng, B.~Yang, T.~Xiao, C.~Zhang, Z.~Wang, R.~Wang,
  and X.~Wang, ``{T-CNN}: Tubelets with convolutional neural networks for
  object detection from videos,'' \emph{{IEEE Transactions on Circuits and
  Systems for Video Technology}}, vol.~28, pp. 2896--2907, 2017.

\bibitem{15}
P.~Bergmann, T.~Meinhardt, and L.~Leal-Taixe, ``{Tracking without bells and
  whistles},'' in \emph{{IEEE/CVF International Conference on Computer Vision
  (ICCV)}}, 2019, pp. 941--951.

\bibitem{1}
X.~Zhou, V.~Koltun, and P.~Kr{\"a}henb{\"u}hl, ``{Tracking objects as
  points},'' in \emph{European Conference on Computer Vision (ECCV)}.\hskip 1em
  plus 0.5em minus 0.4em\relax Springer, 2020, pp. 474--490.

\bibitem{16}
S.~A. Ahmadi, A.~Ghorbanian, and A.~Mohammadzadeh, ``Moving vehicle detection,
  tracking and traffic parameter estimation from a satellite video: A
  perspective on a smarter city,'' \emph{{International Journal of Remote
  Sensing}}, vol.~40, no.~22, pp. 8379--8394, 2019.

\bibitem{18}
A.~Xu, J.~Wu, G.~Zhang, S.~Pan, T.~Wang, Y.~Jang, and X.~Shen, ``{Motion
  detection in satellite video},'' \emph{{Journal of Remote Sensing \& GIS}},
  vol.~6, no.~2, pp. 1--9, 2017.

\bibitem{19}
T.~Yang, X.~Wang, B.~Yao, J.~Li, Y.~Zhang, Z.~He, and W.~Duan, ``{Small moving
  vehicle detection in a satellite video of an urban area},'' \emph{{Journal of
  Sensors}}, vol.~16, no.~9, p. 1528, 2016.

\bibitem{20}
X.~Zhang and J.~Xiang, ``{Moving object detection in video satellite image
  based on deep learning},'' in \emph{{LIDAR Imaging Detection and Target
  Recognition}}, vol. 10605.\hskip 1em plus 0.5em minus 0.4em\relax
  {International Society for Optics and Photonics}, 2017, p. 106054H.

\bibitem{32}
V.~Reilly, H.~Idrees, and M.~Shah, ``{Detection and tracking of large number of
  targets in wide area surveillance},'' in \emph{{European Conference on
  Computer Vision (ECCV)}}.\hskip 1em plus 0.5em minus 0.4em\relax Springer,
  2010, pp. 186--199.

\bibitem{21}
H.~Li, L.~Chen, F.~Li, and M.~Huang, ``{Ship detection and tracking method for
  satellite video based on multiscale saliency and surrounding contrast
  analysis},'' \emph{{Journal of Applied Remote Sensing}}, vol.~13, no.~2, p.
  026511, 2019.

\bibitem{22}
W.~Ao, Y.~Fu, X.~Hou, and F.~Xu, ``{Needles in a haystack: Tracking city-scale
  moving vehicles from continuously moving satellite},'' \emph{{IEEE
  Transactions on Image Processing}}, vol.~29, pp. 1944--1957, 2019.

\bibitem{saleemi}
I.~Saleemi and M.~Shah, ``Multiframe many--many point correspondence for
  vehicle tracking in high density wide area aerial videos,''
  \emph{{International Journal of Computer Vision}}, vol. 104, pp. 198--219,
  2013.

\bibitem{xiao}
J.~Xiao, H.~Cheng, H.~Sawhney, and F.~Han, ``{Vehicle detection and tracking in
  wide field-of-view aerial video},'' in \emph{{IEEE Conference on Computer
  Vision and Pattern Recognition (CVPR)}}, 2010, pp. 679--684.

\bibitem{keck}
M.~Keck, L.~Galup, and C.~Stauffer, ``{Real-time tracking of low-resolution
  vehicles for wide-area persistent surveillance},'' in \emph{IEEE Workshop on
  Applications of Computer Vision}, 2013, pp. 441--448.

\bibitem{24}
Y.~Zhou and S.~Maskell, ``{Detecting and tracking small moving objects in wide
  area motion imagery (WAMI) using convolutional neural networks (CNNs)},'' in
  \emph{{IEEE International Conference on Information Fusion (FUSION)}}, 2019,
  pp. 1--8.

\bibitem{26}
S.~Ji, W.~Xu, M.~Yang, and K.~Yu, ``{3D convolutional neural networks for human
  action recognition},'' \emph{{IEEE Transactions on Pattern Analysis and
  Machine Intelligence}}, vol.~35, no.~1, pp. 221--231, 2012.

\bibitem{27}
K.~Simonyan and A.~Zisserman, ``{Two-Stream} convolutional networks for action
  recognition in videos,'' in \emph{{Advances in Neural Information Processing
  Systems}}, 2014, pp. 568--576.

\bibitem{28}
M.~Baccouche, F.~Mamalet, C.~Wolf, C.~Garcia, and A.~Baskurt, ``{Sequential
  deep learning for human action recognition},'' in \emph{{International
  Workshop on Human Behavior Understanding}}.\hskip 1em plus 0.5em minus
  0.4em\relax Springer, 2011, pp. 29--39.

\bibitem{29}
K.~Kang, W.~Ouyang, H.~Li, and X.~Wang, ``{Object detection from video tubelets
  with convolutional neural networks},'' in \emph{{IEEE Conference on Computer
  Vision and Pattern Recognition (CVPR)}}, 2016, pp. 817--825.

\bibitem{37}
X.~Li, Z.~Jie, W.~Wang, C.~Liu, J.~Yang, X.~Shen, Z.~Lin, Q.~Chen, S.~Yan, and
  J.~Feng, ``{FoveaNet: Perspective-aware urban scene parsing},'' in
  \emph{{IEEE International Conference on Computer Vision (ICCV)}}, 2017, pp.
  784--792.

\bibitem{44}
K.~Duan, S.~Bai, L.~Xie, H.~Qi, Q.~Huang, and Q.~Tian, ``{Centernet: Keypoint
  triplets for object detection},'' in \emph{{IEEE/CVF International Conference
  on Computer Vision (ICCV)}}, 2019, pp. 6569--6578.

\bibitem{42}
H.~Law and J.~Deng, ``{Cornernet: Detecting objects as paired keypoints},'' in
  \emph{{European Conference on Computer Vision (ECCV)}}, 2018, pp. 734--750.

\bibitem{48}
G.~A. Mills-Tettey, A.~Stentz, and M.~B. Dias, ``The dynamic hungarian
  algorithm for the assignment problem with changing costs,'' \emph{{Robotics
  Institute, Pittsburgh, PA, Tech. Rep. CMU-RI-TR-07-27}}, 2007.

\bibitem{45}
O.~Ronneberger, P.~Fischer, and T.~Brox, ``{U-Net: Convolutional networks for
  biomedical image segmentation},'' in \emph{{International Conference on
  Medical Image Computing and Computer-Assisted Intervention}}.\hskip 1em plus
  0.5em minus 0.4em\relax Springer, 2015, pp. 234--241.

\bibitem{47}
A.~Newell, K.~Yang, and J.~Deng, ``{Stacked Hourglass Networks for Human Pose
  Estimation},'' in \emph{{European Conference on Computer Vision
  (ECCV)}}.\hskip 1em plus 0.5em minus 0.4em\relax Springer, 2016, pp.
  483--499.

\bibitem{46}
K.~He, X.~Zhang, S.~Ren, and J.~Sun, ``{Deep Residual Learning for Image
  Recognition},'' in \emph{{IEEE Conference on Computer Vision and Pattern
  Recognition (CVPR)}}, 2016, pp. 770--778.

\bibitem{43}
T.-Y. Lin, P.~Goyal, R.~Girshick, K.~He, and P.~Doll{\'a}r, ``Focal loss for
  dense object detection,'' in \emph{{IEEE International Conference on Computer
  Vision (ICCV)}}, 2017, pp. 2980--2988.

\bibitem{visD}
P.~Zhu, L.~Wen, D.~Du, X.~Bian, Q.~Hu, and H.~Ling, ``{Vision Meets Drones:
  Past, Present and Future},'' 2020.

\bibitem{38}
H.~Pirsiavash, D.~Ramanan, and C.~C. Fowlkes, ``{Globally-optimal greedy
  algorithms for tracking a variable number of objects},'' in \emph{IEEE
  Conference on Computer Vision and Pattern Recognition (CVPR)}, 2011, pp.
  1201--1208.

\bibitem{34}
L.~W. Sommer, M.~Teutsch, T.~Schuchert, and J.~Beyerer, ``{A survey on moving
  object detection for wide area motion imagery},'' in \emph{IEEE Winter
  Conference on Applications of Computer Vision (WACV)}, 2016, pp. 1--9.

\end{thebibliography}

\end{document}